\documentclass[lettersize,journal]{IEEEtran}
\usepackage{amsmath,amsfonts}
\usepackage{algorithmic}
\usepackage{array}
\usepackage{cite}
\usepackage{pdfpages}
\usepackage{tabularx}
\usepackage{threeparttable}
\usepackage{multirow}
\usepackage[caption=false,font=normalsize,labelfont=sf,textfont=sf]{subfig}
\usepackage{textcomp}
\usepackage{stfloats}
\usepackage{url}
\usepackage{verbatim}
\usepackage{graphicx}
\usepackage{hyperref}
\usepackage{makecell}
\usepackage{balance}
\usepackage{soul}
\usepackage{times}
\usepackage{epsfig}
\usepackage{graphicx}
\usepackage{amsmath}
\usepackage{amssymb}
\usepackage{booktabs}
\usepackage{bbding}
\usepackage{multirow}
\usepackage[normalem]{ulem}
\useunder{\uline}{\ul}{}
\usepackage{float}
\usepackage{bm}
\usepackage[table,xcdraw]{xcolor}

\hypersetup{
    hidelinks,
    colorlinks=true,
    allcolors=blue,
    pdfstartview=Fit,
    breaklinks=true
}

\newcommand{\Rmnum}[1]

\def\BibTeX{{\rm B\kern-.05em{\sc i\kern-.025em b}\kern-.08em
    T\kern-.1667em\lower.7ex\hbox{E}\kern-.125emX}}

\title{UV-M$^3$TL: A Unified and Versatile Multimodal Multi-Task Learning Framework for Assistive Driving Perception}
\author{
  Wenzhuo Liu \textit{Graduate Student Member, IEEE}, Qiannan Guo, Zhen Wang, Wenshuo Wang*, \textit{Member, IEEE}, Lei Yang , \textit{Member, IEEE}, Yicheng Qiao, Lening Wang, Zhiwei Li, Chen Lv, \textit{Senior Member, IEEE}, Shanghang Zhang, \textit{Member, IEEE}, Junqiang Xi, \textit{Member, IEEE}, and Huaping Liu, \textit{Fellow, IEEE}
  \thanks{This work was supported by the National Natural Science Foundation of China (Grant No. 52572469). Corresponding author: Wenshuo Wang.}
  \thanks{
    Wenzhuo Liu, Zhen Wang, Wenshuo Wang and Junqiang Xi are with Energy and Transportation Domain, Beijing Institute of Technology, Zhuhai, China, 519088 (e-mail: wzliu@bit.edu.cn; zwang@bit.edu.cn; ws.wang@bit.edu.cn; xijunqiang@bit.edu.cn). 

    Qiannan Guo, Yicheng Qiao and Huaping Liu with the State Key Laboratory of Intelligent Technology and  Systems and Department of Computer Science and Technology, Tsinghua University, Beijing, China, 100084 (e-mail: guoqiannan1203@163.com; yichengqiao21@gmail.com; hpliu@tsinghua.edu.cn).

    Lei Yang and Chen Lv are with the School of Mechanical and Aerospace Engineering, Nanyang Technological University, Singapore (e-mail:lei.yang@ntu.edu.sg; lyuchen@ntu.edu.sg)
    
    Lening Wang is with the School of Transportation Science and Engineering and the State Key Lab of Intelligent Transportation System, Beihang University, Beijing, China, 100191 (e-mail: leningwang@buaa.edu.cn ). 

    Zhiwei Li is with Beijing University of Chemical Technology, Beijing, 100029, China (e-mail: 2022500066@buct.edu.cn).

    Shanghang Zhang is with the State Key Laboratory of Multimedia Information Processing, School of Computer Science, Peking University, Beijing, China, 100871 (e-mail: shanghang@pku.edu.cn).
  }
}
\begin{document}
\maketitle

\begin{abstract}
Advanced Driver Assistance Systems (ADAS) need to understand human driver behavior while perceiving their navigation context, but jointly learning these heterogeneous tasks would cause inter-task negative transfer and impair system performance. Here, we propose a Unified and Versatile Multimodal Multi-Task Learning (UV-M$^3$TL) framework to simultaneously recognize driver behavior, driver emotion, vehicle behavior, and traffic context, while mitigating inter-task negative transfer. Our framework incorporates two core components: dual-branch spatial channel multimodal embedding (DB-SCME) and adaptive feature-decoupled multi-task loss (AFD-Loss). DB-SCME enhances cross-task knowledge transfer while mitigating task conflicts by employing a dual-branch structure to explicitly model salient task-shared and task-specific features. AFD-Loss improves the stability of joint optimization while guiding the model to learn diverse multi-task representations by introducing an adaptive weighting mechanism based on learning dynamics and feature decoupling constraints. We evaluate our method on the AIDE dataset, and the experimental results demonstrate that UV-M$^3$TL achieves state-of-the-art performance across all four tasks. To further prove the versatility, we evaluate UV-M$^3$TL on additional public multi-task perception benchmarks (BDD100K, CityScapes, NYUD-v2, and PASCAL-Context), where it consistently delivers strong performance across diverse task combinations, attaining state-of-the-art results on most tasks.
\end{abstract}


\section{Introduction}
\label{sec:intro}

\begin{figure}
\centering
\includegraphics[width=0.49\textwidth]{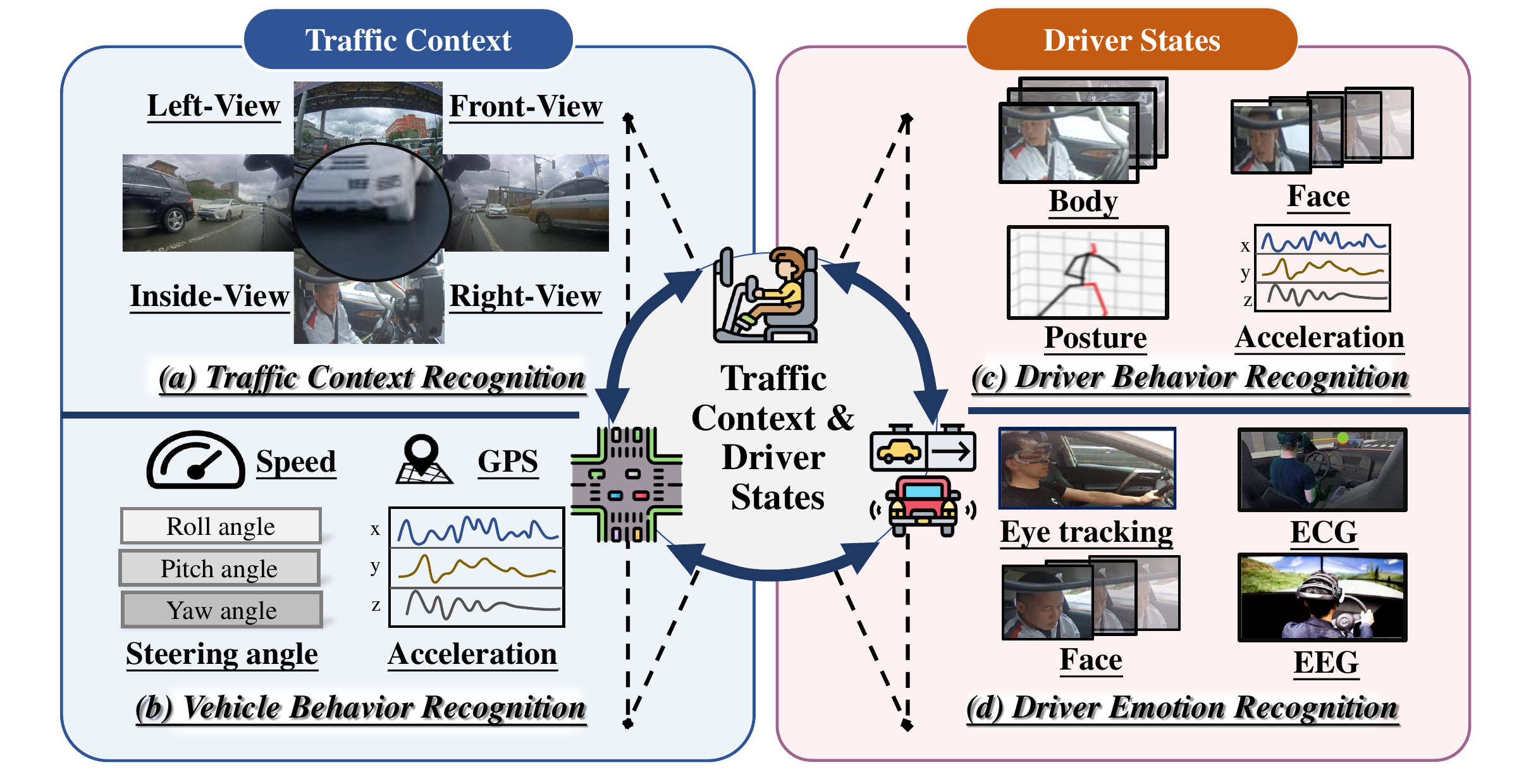}
\vspace{-1em}
\caption{Traffic context and driver states interaction. Tasks (a), (b), (c), and (d) represent traffic context recognition, vehicle behavior recognition, driver behavior recognition, and driver emotion recognition, respectively. These tasks comprehensively demonstrate the complex and closely interconnected relationships between the driver and traffic.}
\label{network1}
\vspace{-1em}
\end{figure}

\IEEEPARstart{O}{ver} the past decade, driving safety and traffic efficiency have been improved through Advanced Driver Assistance Systems (ADAS), such as automatic emergency braking and lane keeping assist \cite{zhang2023oblique,li2024mipdmultisensoryinteractiveperception}. Approximately 1.35 million fatalities occur annually in traffic accidents \cite{world2019global}, of which human drivers' abnormal mental or physical states contributing to over 65\% \cite{tian2013studying}. Accurate driver state identification is a core function of ADAS \cite{martin2019drive} but challenging due to their intricate causal interplays with traffic context \cite{yang2023aide} (see Fig. \ref{network1}). For instance, traffic congestion could induce driver anxiety and affect driving behavior \cite{liu2025tem, yang2023aide}.

Most existing works recognize driver state or traffic context by learning individual tasks independently, such as driver behavior or emotion recognition \cite{saleh2017driving,gong2023sifdrivenet,li2021cogemonet,yang2023robust} and traffic environment recognition \cite{guo2023temporal,martin2018dynamics}, termed single-task learning (STL). However, the STL trains a separate neural network for each task and fails to exploit task interaction, limiting the learning potentials across tasks \cite{vandenhende2021multi, qian2019dlt}. In practice, these tasks are typically linked to each other \cite{yang2023aide,martin2019drive}; for example, a lane-change assistance system should recognize the current traffic context and the driver's cognitive state (e.g., intention, emotion) \cite{guo2023temporal, xing2021multi}.

\begin{figure*}[h]
    \centering
    \includegraphics[width=0.98\textwidth]{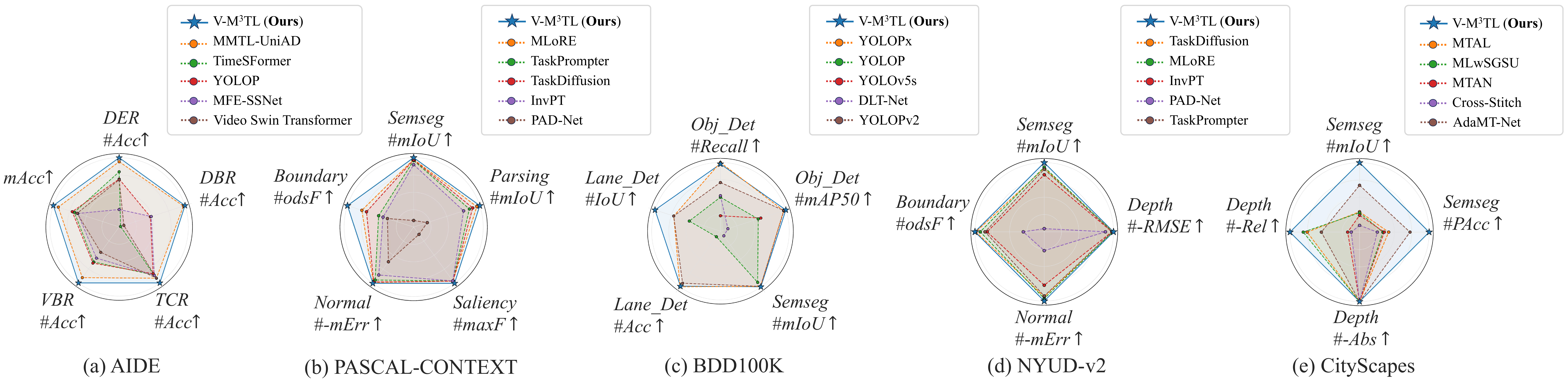}
    \caption{Comparison results of different methods on five benchmark datasets. (a)–(e) present the performance comparisons between our method and various state-of-the-art approaches on the AIDE, PASCAL-Context, BDD100K, NYUD-v2, and CityScapes datasets, respectively. To facilitate a more intuitive comparison of performance across different methods, for evaluation metrics where lower values indicate better performance—such as mErr, RMSE, Absolute Error (Abs), and Relative Error (Rel)—we take their negative values (denoted by “-” in the figure), so that all metrics follow a unified convention where higher values correspond to better performance. Detailed Comparison results of all methods are reported in Section \ref{Experiments}.}
    \label{fig2_de}
    \vspace{-1em}
\end{figure*}

Learning these tasks jointly with a single model, known as Multi-task learning (MTL), can improve task recognition accuracy and generalization through shared representations \cite{zhang2018overview,ishihara2021multi}. However, this approach is often hampered by inter-task negative transfer, where one or more tasks underperform their single-task baselines due to conflicting feature learning across tasks, often arising from divergent predictive targets or input data distributions \cite{vandenhende2021multi}. Many works focus on closely related tasks to mitigate this issue. For instance, in traffic scene understanding, depth estimation and semantic segmentation are often co-modeled by leveraging their geometric–semantic consistency\cite{liu2019end, jha2021s, jha2020adamt, chen2023umt}. Similarly, lane-detection and drivable-area segmentation benefit from spatial prior (lane markings bordering drivable regions). Driver-centric tasks such as behavior/emotion/intention recognition are also jointly learned by exploiting cognitive-behavioral correlations \cite{xing2021multi,luvizon2020multi, xun2020multitask}. Yet, the substantial disparities in both predictive targets and input data between driver-related and environment-related tasks typically induce severe negative transfer. This limitation explains why existing work seldom bridges these two task categories, ultimately restricting ADAS from holistically interpreting driving context \cite{yang2023aide, martin2019drive}. 

In addition, MTL models face two critical challenges. (1) \textbf{Optimization Imbalance}: The loss functions of different tasks often diverge in convergence speed and numerical scale due to distinct learning objectives. These disparities may lead to conflicting gradient direction and magnitude during backpropagation, hindering the model's ability to balance multiple tasks and degrading overall performance \cite{vandenhende2021multi}. To address this issue, current approaches primarily fall into two categories: (i) static weight assignment allocates fixed weights to task losses based on prior knowledge or empirical tuning, addressing scale-induced imbalances \cite{zhao2023drmnet, liang2022effective, miraliev2023real, yang2024multi} and (ii) uncertainty weighting dynamically adjusts loss weights by modeling the aleatoric uncertainty of each task, improving optimization stability \cite{kendall2018multi, zhang2021survey, chen2023umt}. Although these methods show empirical success for specific task combinations, they often overlook intrinsic learning dynamics (e.g., task-specific convergence rates) during training. This limits their versatility across diverse task combinations. Moreover, they focus merely on stabilizing training rather than enabling the model to extract more discriminative features per task, thereby limiting the  MTL model's potential \cite{vandenhende2021multi, zhang2021survey ,shi2023bssnet}. (2) \textbf{Single-modality Limitation}: Many real-world applications inherently demand multimodal data synergy for robust performance \cite{li2021cogemonet, du2020convolution, guo2023temporal, liu2024glmdrivenet, liu2024fmdnet}. For instance, traffic context recognition often requires multi-view images and LiDAR point clouds, while driver behavior/state recognition depends on fused inputs from cabin views, close-up facial images, and skeletal joint data. Although multimodal integration demonstrably improves accuracy \cite{guo2023temporal}, prevailing MTL frameworks predominantly operate on unimodal inputs (e.g., scene images) \cite{wu2022yolop, qian2019dlt, teichmann2018multinet, yang2024multi, zhan2024yolopx, han2022yolopv2}, limiting their real-world applicability.

To address these challenges, we propose UV-M$^3$TL, a unified and versatile multimodal multi-task learning framework that mitigates negative transfer and enables efficient joint optimization across heterogeneous tasks. First, inspired by \cite{wang2020axial,zhu2023biformer}, we design a multi-axis region attention network for multi-view images. This network captures global context via horizontal-vertical attention and then uses region attention to extract interest-triggering features, selecting task-related high-level semantic information. Second, we introduce a dual-branch spatial channel multimodal embedding (DB-SCME) that jointly adjusts task-shared and task-specific parameters across spatial/channel dimensions. This adapts to varying task attention patterns on multimodal inputs, enhancing both specialization and positive transfer. Additionally, we design an adaptive feature-decoupled multi-task loss (AFD-Loss) with dynamic task weighting based on learning progress to alleviate gradient conflicts and feature decoupling constraints that suppress redundancy between task-shared and task-specific features, encouraging diverse representations and synergistic task learning.

We evaluate the proposed method on five multi-task benchmarks covering distinct scenarios: AIDE \cite{yang2023aide}, NYUD-v2 \cite{silberman2012indoor}, PASCAL-Context \cite{chen2014detect}, BDD100K \cite{yu2018bdd100k}, and CityScapes \cite{cordts2016cityscapes}. First, UV-M$^3$TL achieves new state-of-the-art across all four tasks (driver behavior, driver emotion, traffic context, and vehicle behavior recognition)  on the AIDE dataset (see Fig. \ref{fig2_de}  (a)), with an average accuracy gain of 1.41\%-13.50\% over prior works. Ablation studies show that separately learning either driver-related or environment-related tasks causes an accuracy reduction of 3.98\%-5.05\%. Furthermore, our method demonstrates consistent advantages on the remaining four benchmarks which cover dense prediction and panoptic driving perception tasks. It shows excellent performance on all four datasets and achieves state-of-the-art performance on most tasks (see Fig. \ref{fig2_de} (b)-(e)). These results collectively validate UV-M$^3$TL's effectiveness and versatility in complex multi-task scenarios.

In summary, our contributions are fourfold:

\begin{itemize}
    \item [$\bullet$] We propose UV-M$^3$TL, a unified and versatile multimodal multi-task learning framework. By effectively alleviating negative transfer, it enables accurate and stable learning across heterogeneous multimodal inputs and diverse task combinations.
    \item [$\bullet$] We introduce a DB-SCME to extract task-shared and task-specific features, mitigating negative transfer while enhancing cross-task knowledge learning.
    \item [$\bullet$] We design an AFD-Loss to guide the model toward learning diverse and effective representations for different tasks, while enhancing the stability of joint optimization across heterogeneous tasks.
    \item [$\bullet$] To the best of our knowledge, this is the first work to demonstrate strong and consistent performance on five challenging multi-task benchmarks spanning multiple scenarios, validating both the effectiveness of the proposed method and its versatility to diverse multi-task scenarios.
\end{itemize}

A preliminary version of this work was published at 2025CVPR \cite{liu2025mmtl}. This paper extends the conference version with improvements in three main areas. First, we propose a novel dual-branch spatial channel multimodal embedding (DB-SCME) that efficiently extracts task-shared and task-specific features across spatial and channel dimensions. Second, we introduce an adaptive feature-decoupled multi-task loss (AFD-Loss), which incorporates a learning dynamics-based adaptive weighting mechanism and a feature decoupling constraint. This design not only encourages the model to learn diverse features but also enhances the stability of joint optimization, thereby improving the model's prediction accuracy across task combinations. Third, we substantially expand the experimental validation to enable a comprehensive evaluation of UV-M$^3$TL across diverse perception scenarios and task combinations: beyond the coarse perception dataset (AIDE), we further evaluate it on four fine perception benchmarks (NYUD-v2, PASCAL-Context, BDD100K, and CityScapes). Among them, BDD100K and CityScapes are widely used multi-task benchmark datasets in assisted driving perception, while NYUD-v2 and PASCAL-Context encompass richer object categories in indoor and outdoor scenes, presenting greater challenges. The results show that UV-M$^3$TL outperforms MMTL-UniAD across all four tasks on AIDE and also demonstrates excellent performance in   remaining fine perception multi-task scenarios, proving the model's effectiveness and versatility.

\section{Related Work}
\label{sec:Related Work}

\subsection{Multi-task Learning}
\label{Multi-task Learning}

Recent advancements in deep neural networks have made it feasible to learn multiple tasks jointly, improving the performance of individual tasks through shared representations. This process involves two key steps: modeling approach and joint optimization.

\textit{a) Modeling approach:} A key strategy for designing MTL models is the sharing of parameters across tasks. Two primary approaches for parameter sharing are hard parameter sharing and soft parameter sharing. 
\begin{itemize}
    \item Hard parameter sharing consists of a shared encoder and task-specific output heads. By sharing all parameters within the encoder, it reduces overfitting between tasks and enhances learning efficiency.\cite{li2020knowledge,xu2018pad,cao2023relational,al2022zero,huang2024mfe}. However, excessive sharing makes performance highly dependent on inter-task correlations. If tasks have divergent objectives or inherent conflicts, shared a large number of parameters may degrade individual task performance \cite{misra2016cross}.
    \item Soft parameter sharing is more diverse, as it can retain task-specific information within the shared encoder and also facilitate inter-task interaction in task-specific output heads \cite{chen2023adamv,xu2023demt,misra2016cross,yang2024multi,lin2025mtmamba++,ye2022taskprompter}. It mitigates inter-task conflicts even with weak correlations, owing to its flexible parameter sharing strategy. This adaptability enhances each task performance, especially when tasks vary in complexity or semantic relevance.
\end{itemize}

In the context of MTL for ADAS, soft parameter sharing is a promising approach. It enhances performance across multiple driving-related tasks through shared representations while maintaining flexibility in task-specific learning.

\textit{b) Joint optimization:} When multi-task learning objectives exhibit significant divergence, an effective optimization strategy is essential to stabilize training, enhance cross-task synergies \cite{zhang2021survey, zhang2018overview}. Existing multi-task joint optimization strategies in computer vision and ADAS can typically fall into two categories.

\begin{itemize}
    \item Static weight assignment alleviates optimization imbalance caused by heterogeneous task losses by assigning fixed weights to adjust their relative contributions \cite{zhao2023drmnet,zhan2024yolopx,yang2024multi}. While widely adopted in stable task settings \cite{liang2022effective, miraliev2023real}, this heuristic strategy heavily relies on empirical tuning.
    \item Uncertainty weighting leverages aleatoric uncertainty (inherent noise in observation/label) to adaptively adjust the contribution of each task to the overall loss, improving joint optimization stability \cite{zhang2021survey, kendall2018multi, chen2023umt, liu2019end}.
\end{itemize}

Although their successes, they largely ignore the evolving learning dynamics of individual tasks during joint training, restricting their applicability to highly divergent task sets. Moreover, current methods primarily treat the loss function as mere supervisory signals, overlooking their demonstrated potential in single-task learning for guiding feature representation and regularizing model behaviors.

\subsection{Driver State Recognition}

Driver state recognition aims to assess the mental and physical state of the driver, which is essential for ADAS. Existing studies have leveraged various data sources, including driver-specific and traffic-related signals, to infer driver states. 

One line of research relies on vehicle dynamics (e.g., speed, steering angle) complemented by multi-view driving scene images (e.g., front-view, right-view, and left-view) to infer the driver's state \cite{saleh2017driving,gong2023sifdrivenet,liu2024fmdnet}. For example, Liu et al. \cite{liu2024glmdrivenet} designed a global–local attention mechanism that fuses front-view driving scene images with vehicle speed information for driving behavior recognition. While these methods effectively model vehicle-environment interactions, they typically overlook the driver's cognitive or emotional state.

Another studies leverage direct driver monitoring, such as facial images and physiological signals (e.g., heart rate, eye gaze), to detect states like fatigue, stress, or distraction. \cite{du2020convolution,li2021cogemonet,yang2023robust}. For example, Li et al. \cite{li2021cogemonet} employed convolutional neural networks to extract drivers' facial features and jointly modeled them with personalized characteristics (e.g., age, gender, and driving experience) to recognize driver emotions. Chen et al. \cite{chen2021fine} proposed a deep convolutional neural network by integrating driver facial images, eye-tracking data, and multiple physiological signals like electrodermal activity (EDA) to identify driver distraction. However, these methods prioritize driver-specific data and often disregard the influence of real-time traffic conditions on driver behavior.

\begin{figure*}[t]
    \centering
    \includegraphics[width=0.98\textwidth]{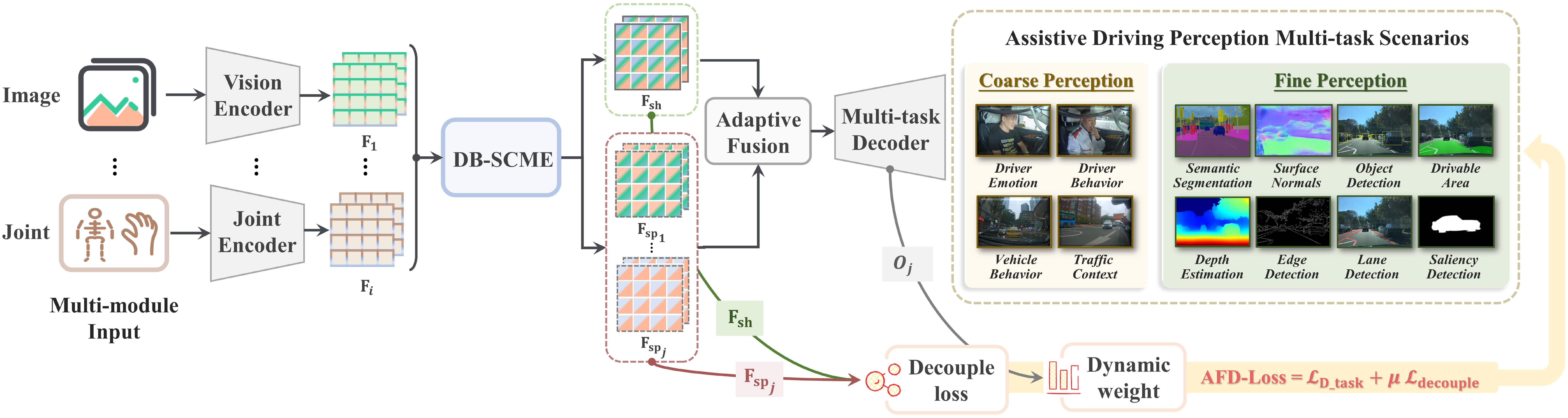}
    \vspace{-1em}
    \caption{The overall pipeline of UV-M$^3$TL with three primary components: Multimodal Encoder, DB-SCME, and AFD-Loss.}
    \label{fig:network-overview}
\end{figure*}

\begin{figure}
\centering
\includegraphics[width=0.48\textwidth]{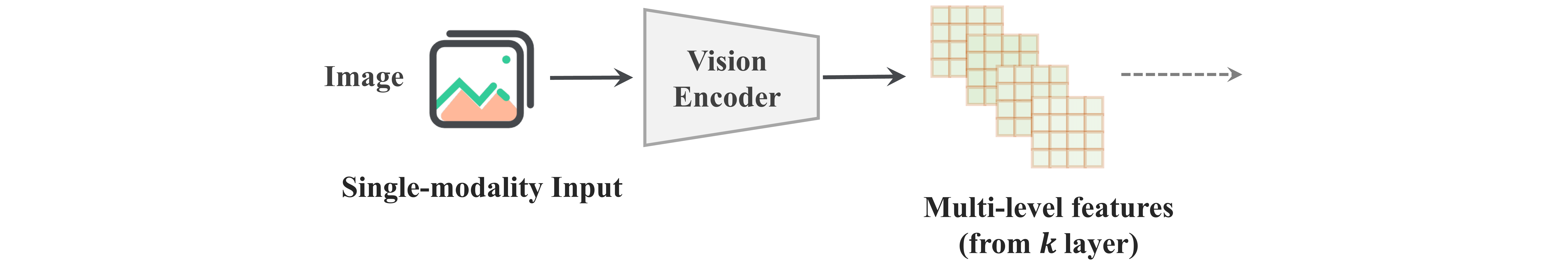}
\vspace{-1em}
\caption{Encoder architecture of UV-M$^3$TL with the single-modality input, where multi-level features are extracted from different network layers.}
\vspace{-1em}
\label{fig:network-overview2}
\end{figure}
Even though their individual merits, existing methods are limited by unilateral focus, either on the driver or the traffic context, failing to integrate both perspectives. This limitation is particularly evident in dynamic driving scenarios, where both the driver's internal state and external traffic context jointly influence driving behavior. Thus, a holistic framework that jointly models driver states and traffic context remains an open challenge in ADAS research, necessitating co-adaptive analysis of both domains.

\begin{figure*}[h]
    \centering
    \includegraphics[width=0.98\textwidth]{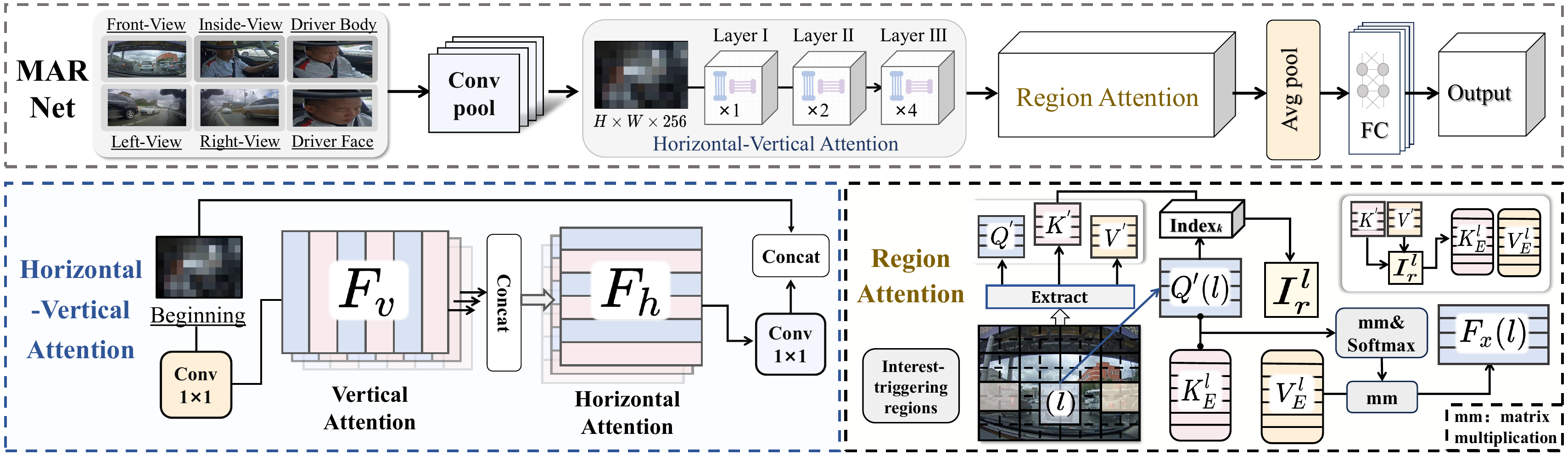}
    \caption{The flowchart of the MARNet architecture, including the processes for horizontal-vertical attention and region attention.}
    \label{fig:GCFANet}
    \vspace{-1em}
\end{figure*}

\begin{figure}
\centering
\includegraphics[width=0.48\textwidth]{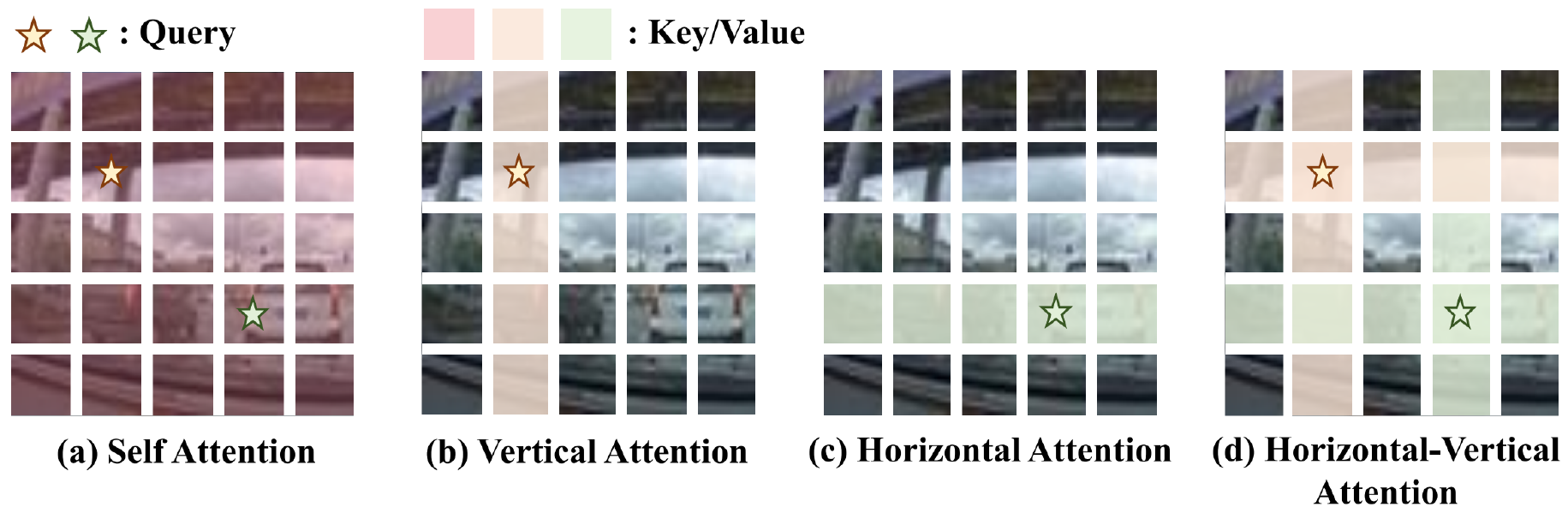}
\vspace{-1em}
\caption{Diagram of different self-attention. (a) represents the most common global self-attention in images; (b) (c) (d) represent vertical attention, horizontal attention, and horizontal-vertical attention, respectively. Among them (d) represents the horizontal-vertical attention we introduced.}
\vspace{-1em}
\label{fig5}
\end{figure}

\section{Methods}

\subsection{Network Overview}
\label{Network Overview}

UV-M$^3$TL consists of three core components (Fig. \ref{fig:network-overview}): a multimodal encoder, DB-SCME, and AFD-Loss. It is worth noting that while UV-M$^3$TL is a multimodal multi-task learning framework, most existing MTL models still rely on single-modality inputs \cite{wu2022yolop,ye2022taskprompter,zhan2024yolopx}. To ensure strong  versatility and efficient joint learning across diverse inputs and  task combinations, UV-M$^3$TL's encoder is designed to support both multimodal and single-modality inputs at the encoder. For multimodal inputs, modality-specific encoders extract features, which are then fed into the DB-SCME for unified modeling (see Fig. \ref{fig:network-overview}). For single-modality input, we follow established practice \cite{chen2024multi,zhan2024yolopx,lin2025mtmamba++} by extracting multi-level features from different encoder layers, which are similarly processed by the DB-SCME (see Fig. \ref{fig:network-overview2}). In the following, we present a detailed description of the overall framework of UV-M$^3$TL, using the multimodal input scenario—the primary focus of this paper—as an illustrative example.

When UV-M$^3$TL uses multimodal inputs, its multimodal encoder integrates a Multi-axis Region Attention Network (MARNet) with a 3D Convolutional Neural Network (3D-CNN). MARNet employs multi-attention mechanisms to extract features from multi-view driving scene images (front-view, right-view, left-view, inside-view, driver face, and driver body), while the 3D-CNN processes kinematic features from driver joint data (gesture and posture). To fully exploit shared and task-specific information, DB-SCME processes these multimodal features through a task-shared branch and a task-specific branch. Specifically, DB-SCME adaptively adjusts the parameters of both branches across spatial and channel dimensions to generate two feature sets: task-shared features $\mathbf{F}_{\mathrm{sh}}$ and task-specific features $\mathbf{F}_{\mathrm{sp}}$. This design enhances cross-task knowledge transfer while capturing task-unique characteristics. For each task $j$ (driver behavior, emotion, traffic context, and vehicle behavior recognition), the final result $O_{j}$ is computed by adaptively fusing the two feature sets and processing them through a task-specific output head:
\begin{equation}
\label{Oj}
O_j = \mathrm{Head}_j (
\sigma(w_j) L^{1}_j(\mathbf{F}_{sh}) + (1 - \sigma(w_j)) L^{2}_j(w_{ca}(\mathbf{F}_{sp_j}))
),
\end{equation}
where $L^{1}_j$ and  $L^{2}_j$  are fully connected layers corresponding to task $j$, $w_{ca}$ denotes a channel attention weighting operation to emphasize task-relevant features, $w_j$ is a learnable weight balancing shared/task-specific contributions, $\sigma$ is the Sigmoid function, and $\mathrm{Head}_j(\cdot)$ denotes the task-specific output head for task $j$. This adaptive fusion emphasizes transferable knowledge across tasks while retaining task-specific discriminability.

Finally, AFD-Loss as the supervisory signal integrates a learning dynamics–based adaptive weighting mechanism and a feature decoupling constraint to enable efficient and stable training of the entire model.

\subsection{Multimodal Encoder}

\subsubsection{Multi-axis Region Attention Network}
\label{Multi-axis Region Attention Network}

Driving scenarios captured from multiple views often include task-unrelated features (e.g., roadside billboards or decorative objects inside the vehicle), which may interfere with multi-task learning. Since the quality of feature extraction critically affects cross-task synergy, selectively sharing task-relevant features enhances inter-task complementarity, whereas irrelevant features can induce negative transfer. To address this issue, we propose MARNet (Fig. \ref{fig:GCFANet}), which leverages horizontal-vertical attention and region attention mechanisms to distill task-discriminative features from multi-view images, thereby reducing negative transfer across tasks. 
  
Let $\mathbf{F}_o \in \mathbb{R}^{H \times W \times C}$ be the input feature map, obtained from the multi-view images through initial convolution operations, where $H$, $W$ and $C$ are the height, width, and number of channels of $\mathbf{F}_o$. The horizontal-vertical attention first performs linear projections on the input feature map using three weight matrices to generate the query, key, and value, denoted as $\mathbf{Q}$, $\mathbf{K}$, and $\mathbf{V}$, respectively. We then apply self-attention along the vertical direction of $\mathbf{F}_o$ at each position ($h,w$) to integrate features relevant (see Fig. \ref{fig5} (b)), generating a new feature $\mathbf{F}_v$ by:
\begin{equation}
\begin{aligned}
\mathbf{F}_v &= \sum_{w=1}^{W} \sum_{h=1}^{H} \sum_{h'=1}^{H} \text{softmax}\left( \frac{ \mathbf{Q}_{(h, w, \cdot)}  \mathbf{K}_{(h', w, \cdot)}^\top }{ \sqrt{C} } \right) \mathbf{V}_{(h', w, \cdot)},
\end{aligned}
\end{equation}
where $\mathbf{Q}_{(h, w, \cdot)}$ denotes the query vector at position $(h, w)$, $\mathbf{K}_{(h', w, \cdot)}$ and $ \mathbf{V}_{(h', w, \cdot)}$ denote the key and value vector at position $(h', w)$, respectively. Similarly, we obtain $\mathbf{F}_h$ along the horizontal direction (see Fig. \ref{fig5} (c)) by:

\begin{equation}
\begin{aligned}
\mathbf{F}_h &= \sum_{h=1}^{H} \sum_{w=1}^{W}  \sum_{w'=1}^{W} \text{softmax}\left( \frac{ \mathbf{F}_{v_{(h, w, \cdot)}}  \mathbf{K}_{(h, w', \cdot)}^\top }{ \sqrt{C} } \right) \mathbf{V}_{(h, w', \cdot)}.
\end{aligned}
\end{equation}

Next, we concatenate features $\mathbf{F}_h$ and $\mathbf{F}_o$ to obtain a new feature map $\mathbf{F}’ \in \mathbb{R}^{H \times W \times C}$. 
\begin{equation}
\mathbf{F}' = \mathrm{Concat}(\mathrm{C2D_{1*1}}(\mathbf{F}_h), \mathbf{F}_o),
\end{equation} 
where $\mathrm{C2D_{1*1}}(\cdot)$ denotes a 2D convolution operation  with kernel size of 1, and $\mathrm{Concat}(\cdot)$ denotes a concatenate operation. This operation utilizes the long-range dependencies extracted along horizontal-vertical directions of $\mathbf{F}_o$ (see Fig. \ref{fig5} (d)), preserving the detailed information of initial features.

Horizontal-vertical attention effectively captures directional features and global context in multi-view images. However, real-world scenarios often involve nearby road users (e.g., other vehicles and pedestrians) with arbitrary physical orientations, which limits the model's ability to capture their spatial dependencies due to the fixed directional bias of horizontal-vertical attention \cite{zhu2023biformer}. To address this limitation, we enhance MARNet with a region attention mechanism. By computing similarity scores across the input feature map, $\mathbf{F}'$, this module dynamically identifies and prioritizes salient regions, adaptively focusing on non-directionally constrained dynamic objects. 

Specifically, we first use region attention to partition the feature map $\mathbf{F}'$ into $HW/t^2$ independent regions, each of size $t \times t$, transforming $\mathbf{F}'$ into $\mathbf{F}'' \in \mathbb{R}^{\frac{HW}{t^2} \times t^2 \times C}$. We then apply three weight matrices to linearly project $\mathbf{F}''$ into the query, key, and value, denoted as $\mathbf{Q}'$, $\mathbf{K}'$, $\mathbf{V}' \in \mathbb{R}^{\frac{HW}{t^2} \times t^2 \times C}$, respectively. To improve computational efficiency, we apply a global average pooling across the second dimension of both $\mathbf{Q}'$ and $\mathbf{K}'$, resulting in $\mathbf{Q}''$, $\mathbf{K}'' \in \mathbb{R}^{\frac{HW}{t^2} \times C}$. The similarity matrix is then computed using the dot product between $\mathbf{Q}''$ and $\mathbf{K}''$, and the $k$ most similar regions for each region $l$ are selected, forming the index set $\mathbf{I}_r^l \in \mathbb{R}^{k}$:
\begin{equation}
\mathbf{I}_r^l = \text{Index}_k\left( \mathbf{Q}''(l) \mathbf{K}''^\top \right ),
\end{equation}
where $\mathbf{Q}''(l) \in \mathbb{R}^{1 \times C}$ is the query vector for region $l$, and $\mathbf{K}''^\top \in \mathbb{R}^{C \times \frac{HW}{t^2}}$ is the transpose of $\mathbf{K}''$. With this index, we extract the corresponding rows from $\mathbf{K}'$ and $\mathbf{V}'$ to form $\mathbf{K}_E^l \in \mathbb{R}^{k \times t^2 \times C}$ and $\mathbf{V} _E^l\in\mathbb{R}^{k \times t^2 \times C} $, respectively. Attention is then computed for each region $l$ and its top $k$ most similar regions:
\begin{equation}
\mathbf{F}_x(l) = \text{softmax}\left( \frac{ \mathbf{Q}'(l) (\mathbf{K}_E^l)^\top }{ \sqrt{C} } \right) \mathbf{V}_E^l,
\end{equation}
where $\mathbf{F}_x(l) \in \mathbb{R}^{t^2 \times C}$ is the updated feature for region $l$ after weighting the features via attention. After this operation, the output feature for all regions is $\mathbf{F}_x \in \mathbb{R}^{\frac{HW}{t^2} \times t^2 \times C}$. Finally, we reshape  $\mathbf{F}_x$ back to the dimension $(H, W, C)$, apply global average pooling, and pass the pooled result through a fully connected layer to obtain the final feature representation $\mathbf{F}_i$, where $i \in \{1, 2, 3, 4, 5, 6\}$ corresponds to the six multi-view image inputs.

MARNet extracts deep features from multi-view images by combining horizontal-vertical attention with region attention, yielding a rich set of effective features for subsequent multimodal feature embedding.

\subsubsection{3D-CNN}

In the multimodal encoder, we use a 3D-CNN to analyze the driver's gestures and postures from continuous video frames by applying 3D convolutional kernels across temporal and spatial dimensions. The feature representation is denoted as $\mathbf{F}_i$, where $i \in \{7,8\}$ represents two joint inputs. The extracted features are used to understand driver behavior and emotion.

\begin{figure*}
\centering
\includegraphics[width=0.95\textwidth]{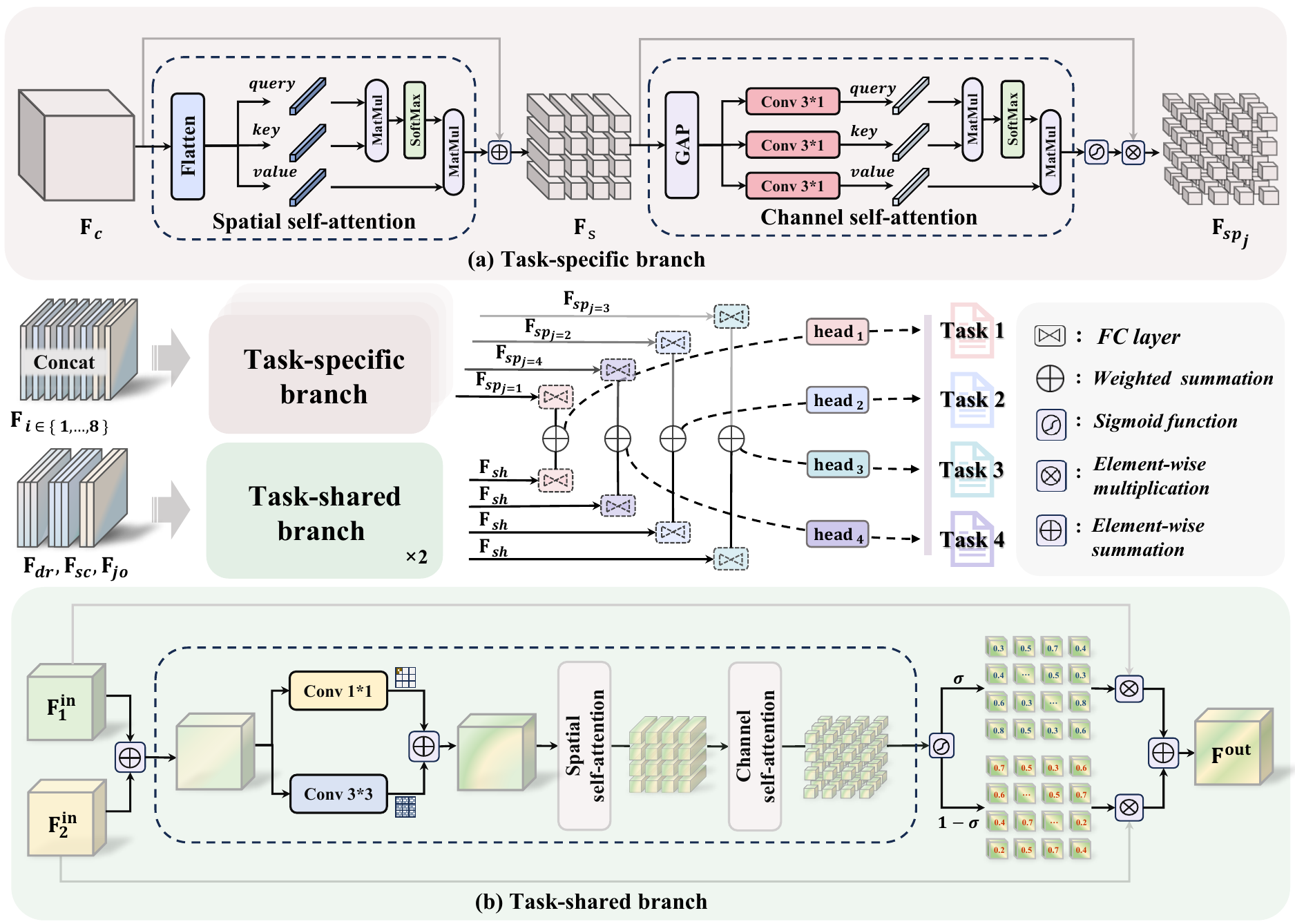}
\vspace{-1em}
\caption{Algorithm flowchart of DB-SCME with (a) a task-specific branch and (b) a task-shared branch.}
\vspace{-1em}
\label{SME}
\end{figure*}

\subsection{DB-SCME}
\label{Dual-Branch Spatial–channel Multimodal Embedding}
In MTL, balancing shared and task-specific features is crucial. Task-shared features facilitate knowledge transfer across tasks, improving model generalization \cite{cao2023relational,al2022zero}. However, excessive sharing may fail to capture task disparities, leading to negative transfer. Task-specific features help mitigate task conflict and reduce the risk of negative transfer. Yet, over-reliance on these features can limit information sharing, degrading cross-task generalization \cite{chen2023adamv,xu2023demt}. To address this issue, we propose DB-SCME (Fig.\ref{SME}), a dual-branch architecture that simultaneously extracts both feature types and dynamically balances their contributions based on the specific tasks.

The DB-SCME has two main components: a task-shared branch and a task-specific branch. Negative transfer occurs because different tasks require distinct information from the same data \cite{vandenhende2021multi}, such as object locations and semantic cues. These discrepancies are typically encoded in the spatial and channel dimensions of features \cite{olah2017feature}. The task-specific branch uses two modules: spatial self-attention and channel self-attention. These model multimodal inputs spatially and channel-wise, extracting task-specific features and reducing negative transfer (Fig.\ref{SME}(a)). First, the spatial self-attention module flattens the feature $\mathbf{F}_c$ along the spatial dimensions. Given $\mathbf{F}_c \in \mathbb{R}^{H' \times W' \times C'}$, we flattened it spatially to $(H'W', C')$, where each $(1, C')$ vector represents a specific location. Here, $\mathbf{F}_c$ is constructed by concatenating multimodal features ${\mathbf{F}_i}$ along the channel dimension, i.e., $\mathbf{F}_c = \mathrm{Concat}(\mathbf{F}_1, \mathbf{F}_2, \ldots, \mathbf{F}_8)$. The flattened features are then linearly projected into query, key, and value, and a multi-head attention mechanism is employed to model the relationships among different spatial locations, thereby guiding the model to focus on regions relevant to the current task's objective. Finally, the attention-enhanced features are reshaped back to the original spatial resolution $(H', W', C')$ and combined with the feature $\mathbf{F}_c$ via a residual connection to produce the output feature $\mathbf{F}_s$, which is computed as follows:
\begin{equation}
\label{Fs}
        \mathbf{F}_s = \mathbf{F}_c + \phi^{-1}(\mathrm{{MHA}}_j(\mathbf{L}(\phi((\mathbf{F}_c)),n)),
\end{equation}
where $\mathrm{MHA}_j(a,b)$ is the multi-head self-attention for task $j$, $\mathbf{L}(\cdot)$ denotes the linear projection operation, $n$ is the number of heads, and $\phi(\cdot)$ and $\phi^{-1}(\cdot)$ represent the spatial flattening operation and its inverse that reshape features between $(H', W', C')$ and $(H'W', C')$, respectively.

Following the spatial self-attention module, the channel self-attention module applies global average pooling to the feature $\mathbf{F}_s$. Then, three 1D convolutions (kernel size = 3) transform these features into query, key, and value vectors, capturing local channel dependencies. The multi-head attention subsequently models global channel interactions. The global features are normalized through a Sigmoid function to $(0,1)$, which then reweights $\mathbf{F}_s$ through element-wise multiplication ($\otimes$). This produces task-specific features $\mathbf{F}_{\mathrm{sp}_j}$ that emphasize relevant channels while suppressing irrelevant ones. This design is particularly effective for multimodal inputs, where $\mathbf{F}_s$ contains distinct modality features (MARNet and 3D-CNN outputs). As shown in Section \ref{Ablation studies on multimodal data}, the module automatically enhances modality-specific features according to task requirements, for instance, prioritizing facial/body images for driver behavior/emotion recognition. The channel interaction dynamically adjusts the importance of each modality and feature representations across different tasks, reducing noise from irrelevant information. The computation process is:
\begin{equation}
\label{sp}
        \mathbf{F}_{\mathrm{sp}_j} = \mathbf{F}_s \otimes \sigma(\mathrm{{MHA}}_j(\mathrm{{C1D}}_j(\mathrm{GAP}(\mathbf{F}_s)), n)) ,
\end{equation}
where $\mathrm{{C1D}}_j(\cdot)$ denotes the 1D convolution. 

\begin{figure*}
\centering
\includegraphics[width=0.95\textwidth]{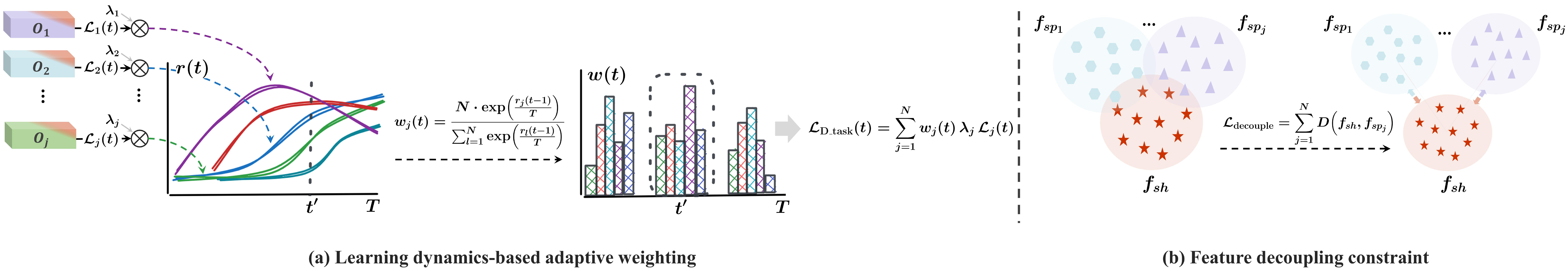}
\vspace{-1em}
\caption{Algorithm flowchart of AFD-Loss. (a) and (b) illustrate the computation processes of $\mathcal{L}_{\mathrm{D\_task}}$ and $\mathcal{L}_{\mathrm{decouple}}$, respectively.}
\vspace{-1em}
\label{AFD}
\end{figure*}

To enhance multi-task learning, we designed a task-shared branch (Fig. \ref{SME} (b)). Multimodal features are categorized by source and structure: traffic context images (i.e., left, right, and front views) form $\mathbf{F}_{\mathrm{sc}}$, driver-related images (i.e., inside view, face, and body) form $\mathbf{F}_{\mathrm{dr}}$, and joint features (i.e., posture and gesture) form $\mathbf{F}_{\mathrm{jo}}$. The task-shared branch combines $\mathbf{F}_{\mathrm{sc}}$ and $\mathbf{F}_{\mathrm{dr}}$, then applies convolutions with kernel sizes of 1 and 3 to capture features at different receptive-field scales, enabling cross-modal semantic alignment. The process $f(\cdot)$ is defined as:
\begin{equation}
f(\mathbf{F}_{\mathrm{dr}},\mathbf{F}_{\mathrm{sc}}) = \mathrm{C2D}_{1*1}(\mathbf{F}_{\mathrm{dr}} + \mathbf{F}_{\mathrm{sc}}) + \mathrm{C2D}_{3*3}(\mathbf{F}_{\mathrm{dr}} + \mathbf{F}_{\mathrm{sc}}),
\end{equation}
where $\mathrm{C2D}_{1*1}$ and $\mathrm{C2D}_{3*3}$ denote 2D convolution with kernel sizes of 1 and 3, respectively. The task-specific branch $\mathbf{T}_{\mathrm{sp}}(\cdot)$ then dynamically integrates this feature map and applies a Sigmoid function to generate the corresponding weights. These weights merge $\mathbf{F}_{\mathrm{sc}}$ and $\mathbf{F}_{\mathrm{dr}}$ into preliminary shared features $\mathbf{F}_{\mathrm{ps}}$:
\begin{align}
    \mathbf{F}_{\mathrm{ps}} = \mathbf{T}_{\mathrm{sh}}(\mathbf{F}_{\mathrm{dr}}, \mathbf{F}_{\mathrm{sc}}) &= \mathbf{F}_{\mathrm{sc}} \times \sigma(\mathbf{T}_{\mathrm{sp}}(f(\mathbf{F}_{\mathrm{dr}} , \mathbf{F}_{\mathrm{sc}})))  \\ 
    & + \mathbf{F}_{\mathrm{dr}} \times (1 - \sigma(\mathbf{T}_{\mathrm{sp}}(f(\mathbf{F}_{\mathrm{dr}} , \mathbf{F}_{\mathrm{sc}}))))\notag .
\end{align}

Similarly, $\mathbf{F}_{\mathrm{ps}}$ and $\mathbf{F}_{\mathrm{jo}}$ are processed by the task-shared branch using the same operations, yielding the task-shared features $\mathbf{F}_{\mathrm{sh}}$. This recursive fusion strategy effectively integrates diverse features from different sources, enabling cross-feature information exchange and enhancing the representational capacity of $\mathbf{F}_{\mathrm{sh}}$. The calculation is:
\begin{equation}
    \mathbf{F}_{\mathrm{sh}} = \mathbf{T}_{\mathrm{sh}}(\mathbf{F}_{\mathrm{jo}}, \mathbf{F}_{\mathrm{ps}}).
\end{equation}

Finally, for task $j$, the task-specific features $\mathbf{F}{\mathrm{sp}_j}$ and the task-shared features $\mathbf{F}_{\mathrm{sh}}$ are dynamically fused and subsequently processed through a task-specific output head via Eq. (\ref{Oj}) to produce the recognition result $O_j$.

\subsection{AFD-Loss}
\label{AFD-Loss}

Heterogeneous tasks often exhibit distinct, time-varying learning dynamics during joint optimization, such as differences in convergence speed. These disparities can cause simpler tasks to converge rapidly in early training stages, producing large gradients that overshadow slower or more complex tasks and destabilize the optimization process. Thus, balancing these dynamics is crucial for stable training and improving the model’s versatility across different task combinations. 

Beyond supervisory signals, loss functions in single-task learning have been shown to guide models towards specific features or behaviors \cite{shi2023bssnet}. However, existing MTL research focuses on architecture design, with less attention to loss functions. Most current methods use static weight assignment (Section \ref{Multi-task Learning}) to address differing loss scales across tasks, achieving stable performance on established task combinations \cite{yang2024multi, zhan2024yolopx, zhao2023drmnet, lin2025mtmamba++}. Yet, such methods fail to adapt to learning dynamics or fully exploit the loss function's potential in facilitating cross-task synergy and mitigating negative transfer.

To address these issues, we propose the AFD-Loss. This loss function consists of two components: $\mathcal{L}_{\mathrm{D\_task}}$ and $\mathcal{L}_{\mathrm{decouple}}$. The overall loss function $\mathcal{L}_{\mathrm{AFD}}$ is defined as follows:
\begin{equation}
\mathcal{L}_{\mathrm{AFD}}
= \mathcal{L}_{\mathrm{D\_task}} + \mu  \, \mathcal{L}_{\mathrm{decouple}},
\end{equation}
where $\mu$ is a hyperparameter that controls the contribution of $\mathcal{L}_{\mathrm{decouple}}$ to the total loss. To address the optimization imbalance caused by tasks with different learning difficulties and convergence rates, we design $\mathcal{L}_{\mathrm{D\_task}}$ based on prior work \cite{liu2019end}. Unlike conventional static weight assignment, $\mathcal{L}_{\mathrm{D\_task}}$ uses a learning dynamics-based adaptive weighting mechanism that dynamically adjusts the learning weights for each task. This is achieved by computing the relative changes in each task's loss between two consecutive training iterations (see Fig. \ref{AFD} (a)).

Let $\mathcal{L}_j(t)$ denote the loss of the $j$-th task at iteration $t$. The relative change rate $r_j(t)$ of the task loss is defined as:
\begin{equation}
r_{j}(t)=\left\{\begin{array}{c}
\frac{\mathcal{L}_{j}(t-1)}{\mathcal{L}_{j}(t-2)}, \mathcal{L}_{j}(t-1)>0 \\
1, \text { otherwise }
\end{array}\right.,
\end{equation}
This ratio measures how quickly the task loss decreases. When $r_{j}(t)$ is large, the loss decreases slowly, implying that the model finds this task harder to learn. When $r_{j}(t)$ is small, the loss drops faster, indicating an easier task. Based on $r_{j}(t)$, we define task weight $w_{j}(t)$ using a softmax function:
\begin{equation}
w_{j}(t) = \frac{N \cdot \exp \left( \frac{r_j(t-1)}{T} \right)}{\sum_{l=1}^{N} \exp \left( \frac{r_l(t-1)}{T} \right)},
\end{equation}
where $T$ is a temperature parameter controlling weight distribution smoothness, and $N$ is the number of tasks. Finally, the dynamic task loss $\mathcal{L}_{\mathrm{D\_task}}$ is:
\begin{equation}
\mathcal{L}_{\mathrm{D\_task}}(t)=\sum_{j=1}^{N} w_{j}(t) \lambda_{j} \mathcal{L}_{j}(t),
\end{equation}
where $\lambda_j$ is the fixed weight for task $j$ to balance loss scales during joint optimization, and $L_{j}(t)$ is the cross-entropy loss of the $j$-th task at iteration $t$.

The information encoded in task-specific and task-shared features is crucial for MTL. If these two features become overly similar, joint learning performance may weaken. Specifically, this similarity causes two issues: (i) Task-specific features can not capture distinct task characteristics, leading to task conflicts, and (ii) shared features can not model cross-task commonalities, reducing the model’s generalization. To enhance cross-task synergy and suppress negative transfer via the loss function, we propose $\mathcal{L}_{\mathrm{decouple}}$. This loss imposes a feature decoupling constraint, encouraging the model to learn more complementary and discriminative shared/specific features while minimizing redundant overlaps between them (see Fig. \ref{AFD} (b)). The loss is defined as:
\begin{equation}
\mathcal{L}_{\mathrm{decouple}}=\sum_{j=1}^{N}D\left ( f_{sh}, f_{sp_j} \right ),
\end{equation}
where $f_{sh}$ and $f_{sp_j}$ denote the task-shared and task-specific features of the $j$-th task, respectively. $D(\cdot,\cdot)$ measures their cosine similarity in the feature space. Minimizing this term suppresses excessive feature correlation, reducing information redundancy.

\section{Experiments}
\label{Experiments}

We evaluate the proposed UV-M$^3$TL on five benchmark datasets (i.e., AIDE \cite{yang2023aide}, BDD100K \cite{yu2018bdd100k}, NYUD-v2 \cite{silberman2012indoor}, PASCAL-Context \cite{chen2014detect}, and CityScapes \cite{cordts2016cityscapes}), covering both coarse and fine perception tasks. Specifically, AIDE consists of four driver- and traffic environment–related classification tasks and is therefore categorized as coarse perception. The remaining datasets involve fine perception tasks such as semantic segmentation, depth estimation, and object detection, and are thus classified as fine perception. Among them, fine perception is further divided according to task characteristics into panoptic driving perception (BDD100K) and dense prediction (NYUD-v2, PASCAL-Context, and CityScapes ). We conduct experiments in the aforementioned diverse perception scenarios to comprehensively validate the effectiveness and versatility of UV-M$^3$TL. 

In this section, we first detail these five datasets (Section \ref{Dataset}) from data modalities, annotation settings, data preprocessing (Section \ref{Data Preprocessing}), evaluation metrics (Section \ref{Evaluation Metrics}), and implementations (Section \ref{Implement Details}). We then compare UV-M$^3$TL with state-of-the-art methods on the coarse perception dataset AIDE  across four tasks (Section \ref{Comparison with the State-of-the-Art}). We further evaluate the versatility of UV-M$^3$TL on four benchmarks in dense prediction and panoptic driving perception (Section \ref{Evaluation on Other Multi-Task Benchmarks}). Finally, we conduct a series of ablation studies to verify the effectiveness of our designed components in UV-M$^3$TL (Section \ref{Ablation Experiment}).

\subsection{Dataset}
\label{Dataset}

\begin{itemize}
\item \textbf{AIDE} \cite{yang2023aide} comprises 2,898 samples with multi-view video data (i.e., front, right, left, inside views), multimodal driver-related data (i.e., face/body images, and skeletal joint coordinates capturing posture/gestures). The dataset is partitioned into training (65\%), validation(15\%), and test (20\%) sets. Each sample is annotated for four  tasks of recognizing driver emotion, driver behavior, traffic context, and vehicle behavior.
\item \textbf{BDD100K} \cite{yu2018bdd100k} contains 100K frames with annotations for 10 tasks with 70K training, 20K validation, and 10K testing samples. The dataset covers urban streets, rural roads, and traffic jam scenes, and diverse weather conditions (sunny, rainy, foggy, and snowy). Since the annotations of the test set are not publicly available, all evaluations in this work are conducted on the validation set. Following \cite{wu2022yolop, zhan2024yolopx}, we select three key panoptic driving perception: traffic object detection, drivable area segmentation, and lane detection.
\item \textbf{NYUD-v2} \cite{silberman2012indoor} contains 1,449 indoor scene images, split into 795 training and 654 testing samples. Each image is annotated with four tasks: 40-class semantic segmentation (Semseg), monocular depth estimation (Depth), surface normal estimation (Normal), and object boundary detection (Boundary).
\item \textbf{PASCAL-Context} \cite{chen2014detect} contains 10,103 images, split into 4,998 training and 5,105 testing samples. Derived from the PASCAL dataset \cite{everingham2010pascal}, this benchmark covers diverse indoor and outdoor scenes, offering high-quality annotations for multiple tasks: semantic segmentation, human parsing (Parsing), saliency detection, surface normal estimation, and object boundary detection.
\item \textbf{Cityscapes} \cite{cordts2016cityscapes} is a dataset of urban street-scene images with high-quality annotations for semantic segmentation and depth estimation. For semantic segmentation, annotations are provided at three granularity levels, covering 2, 7, and 19 classes. Following \cite{liu2019end, chen2023umt}, we use the 7-class setting to ensure fair comparison with existing methods.
\end{itemize}

\subsection{Data Preprocessing}
\label{Data Preprocessing}

\textbf{AIDE}: We preprocess data by following our previous work \cite{liu2025mmtl}. Specifically, we crop the driver's face and body regions from inside-view images using bounding box coordinates. The multi-view video data and skeleton joint data are then synchronized at 16 frames per second for temporal alignment. To maintain sequential information during feature extraction, each model is fed as a sequence of 16 consecutive frames, each frame paired to its corresponding joint data. Finally, we apply data augmentation to all images, including random horizontal and vertical flipping with a 50\% probability.

\textbf{BDD100K}: We follow the common practice \cite{wu2022yolop, han2022yolopv2, zhan2024yolopx}. Specifically, the original image resolution is resized from $1280 \times 720$ to $640 \times 384$. For semantic categories, the classes \textit{car}, \textit{truck}, \textit{bus}, and \textit{train} are merged into a single \textit{vehicle} category, while the \textit{direct} and \textit{alternative} drivable area classes are unified as \textit{drivable}. 

\textbf{NYUD-v2} and \textbf{PASCAL-Context}: We follow the data preprocessing pipeline proposed by \cite{yang2024multi}, including random scaling with scale factors (0.5-2.0), random cropping, horizontal flipping with a probability of 50\%, normalization, and other data augmentation operations.

\textbf{CityScapes}: To accelerate training, all images in both training and validation phases are uniformly resized to $128 \times 256$, while the remaining preprocessing operations align with those in \cite{liu2019end}.

\subsection{Evaluation Metrics}
\label{Evaluation Metrics}

\textbf{AIDE}: Following \cite{fan2022mlfnet,shi2023bssnet}, we introduce a multi-task evaluation metric, referred to as mean accuracy ($\mathrm{mAcc}$, \%):
\begin{equation}
    \mathrm{mAcc} = \frac{1}{N} \sum_{j=1}^{N} \mathrm{Acc}_j,
\end{equation}
where $\mathrm{Acc}_j$ is the accuracy achieved on task $j$. We use mAcc to provide a comprehensive evaluation across all tasks, balancing model performance rather than focusing on individual tasks.

\textbf{BDD100K}: We evaluate traffic object detection using Recall and mAP50, drivable area segmentation using mIoU, and lane detection using Pixel Accuracy (PAcc) and IoU. 

\textbf{NYUD-v2} and \textbf{PASCAL-Context}: We adopt mean Intersection-over-Union (mIoU) to evaluate semantic segmentation and human parsing by following \cite{yang2024multi, lin2025mtmamba++}. Monocular depth estimation is assessed using root mean square error (RMSE), saliency detection is evaluated by the maximum F-measure (maxF), surface normal estimation is measured by the mean angular error (mErr), and object boundary detection is evaluated using the optimal dataset-scale F-measure (odsF).

\textbf{CityScapes}: We evaluate semantic segmentation via mIoU and pixel accuracy and assess depth estimation performance via Absolute Error (Abs.) and Relative Error (Rel.).

\subsection{Implement Details}
\label{Implement Details}
All experiments were conducted using four NVIDIA A40 GPUs. For region attention, we set $t=7$ (region size as $7 \times 7$) and $k=4$ (i.e., selecting the top 4 most similar regions). The multi-head attention module uses $n=8$ heads. For AFD-Loss, the parameter $u$ is initialized to 0.01 and gradually increased during training with the temperature parameter $T=2$.

\textbf{AIDE}: The batch size is 24 for both training and validation. We use the stochastic gradient descent optimizer with a momentum of 0.9 and a weight decay of 0.0001. The initial learning rate is set to 0.1, and the model is trained for 100 epochs. 

\textbf{BDD100K}: Following \cite{zhan2024yolopx}, we employ an AdamW optimizer with momentum parameter (0.937), an initial learning rate of 0.001 along with a cosine annealing schedule, and train for 200 epochs. 

\textbf{NYUD-v2} and \textbf{PASCAL-Context}: The batch size is set to 4, and models are trained for 40,000 iterations. The loss function for each task and the hyperparameter settings follow those used in prior works \cite{yang2024multi, ye2022taskprompter}. 

\textbf{CityScapes}: Training runs for 200 epochs, with a batch size of 2. The optimizer used is Adam, with a learning rate of 0.0001, and StepLR is employed as the learning rate scheduler, which multiplies the learning rate by 0.5 every 100 epochs.

\begin{table*}[t]
\setlength{\tabcolsep}{4pt}
\centering
\renewcommand{\arraystretch}{1.2}
\vspace{-1em}
\caption{Comparison results of baselines on the AIDE dataset for four tasks: Driver Emotion Recognition (DER), Driver Behavior Recognition (DBR), Traffic Context Recognition (TCR), Vehicle Behavior Recognition (VBR). The best results are highlighted in \textbf{bold}, while the second-best results are \underline{underlined}. Scene denotes multi-view images. \textbf{Res}: ResNet~\cite{he2016deep}, \textbf{MLP}: multi-layer perceptron, \textbf{SE}: spatial embedding, \textbf{TE}: temporal embedding, \textbf{TransE}: transformer encoder~\cite{vaswani2017attention}.}
\resizebox{\linewidth}{!}{%
\begin{tabular}{c|cccccl|c|c|c|c|c}
\toprule
 & \multicolumn{6}{c|}{\textbf{Backbone}} & \textbf{DER} & \textbf{DBR} & \textbf{TCR} & \textbf{VBR} &  \\ \cline{2-11}
\multirow{-2}{*}{\centering \textbf{Pattern}} & \textbf{Face} & \textbf{Body} & \textbf{Gesture} & \textbf{Posture} & \multicolumn{2}{c|}{\textbf{Scene}} & \textbf{Acc$\uparrow$} & \textbf{Acc$\uparrow$} & \textbf{Acc$\uparrow$} & \textbf{Acc$\uparrow$} & \multirow{-2}{*}{\centering \textbf{mAcc$\uparrow$}}\\ 
\midrule

\multirow{15}{*}{2D}  & Res18~\cite{he2016deep}         & Res34~\cite{he2016deep}            & MLP+SE  & MLP+SE  & \multicolumn{2}{c|}{PP-Res18~\cite{zhou2017places}}         & 69.05          & 63.87          & 88.01          & 78.16         & 74.77                 \\
 & Res18~\cite{he2016deep}            & Res34~\cite{he2016deep}            & MLP+SE  & MLP+SE  & \multicolumn{2}{c|}{Res34~\cite{he2016deep}}            & 71.26 & 65.35& 83.74          & 77.12          & 74.37               \\
& Res34~\cite{he2016deep}            & Res50~\cite{he2016deep}            & MLP+SE  & MLP+SE  & \multicolumn{2}{c|}{Res50~\cite{he2016deep}}            & 69.68          & 59.77         & 80.13          & 71.26         & 70.21                \\
 & VGG13~\cite{simonyan2014very}            & VGG16~\cite{simonyan2014very}            & MLP+SE  & MLP+SE  & \multicolumn{2}{c|}{VGG16~\cite{simonyan2014very}}            & 70.72          & 63.65          & 82.77          & 77.94         & 73.77                  \\
 & VGG16~\cite{simonyan2014very}            & VGG19~\cite{simonyan2014very}            & MLP+SE  & MLP+SE  & \multicolumn{2}{c|}{VGG19~\cite{simonyan2014very}}            & 69.31          & 62.34          & 83.58          &75.13          & 72.59       \\
 & CPVT~\cite{chu2023conditional}           & CPVT~\cite{chu2023conditional}              & ST-GCN  & ST-GCN  & \multicolumn{2}{c|}{CPVT~\cite{chu2023conditional}}              & 69.01          & 67.35          & 91.44          & 79.57 & 76.84  \\
 & CMT~\cite{Guo_2022_CVPR}           & CMT~\cite{Guo_2022_CVPR}              & ST-GCN  & ST-GCN  & \multicolumn{2}{c|}{CMT~\cite{Guo_2022_CVPR}}              & 68.75          & 68.75          & 93.75          & 81.38 & 78.16  \\
 & GroupMixFormer~\cite{ge2023advancing}           & GroupMixFormer~\cite{ge2023advancing}              & ST-GCN  & ST-GCN  & \multicolumn{2}{c|}{GroupMixFormer~\cite{ge2023advancing}}              & 66.29          & 67.54          & 92.12          & 77.63 & 75.90  
 \\  & AbSViT~\cite{liu2024vision}           & AbSViT~\cite{liu2024vision}              & ST-GCN  & ST-GCN  & \multicolumn{2}{c|}{AbSViT~\cite{liu2024vision}}              & 69.15          & 67.84          & 92.07          & 80.82 & 77.47
 \\  & GLMDriveNet~\cite{liu2024glmdrivenet}             & GLMDriveNet~\cite{liu2024glmdrivenet}              & ST-GCN  & ST-GCN  & \multicolumn{2}{c|}{GLMDriveNet~\cite{liu2024glmdrivenet}}              & 71.38          & 66.57          & 90.23          & 77.19          & 76.34
 \\ & HAT \cite{liu2024vision}             & HAT \cite{liu2024vision}             & ST-GCN  & ST-GCN  & \multicolumn{2}{c|}{HAT \cite{liu2024vision}}              & 75.00          & 70.31          & 92.18          & 81.25          & 79.69    
 \\ & MTS-Mamba \cite{liu2025tem}      & MTS-Mamba \cite{liu2025tem} & 3DCNN & 3DCNN   &  \multicolumn{2}{c|}{MTS-Mamba \cite{liu2025tem}} & 75.00 & 69.31 & \underline{96.29} & \underline{86.11} & 81.68 \\
 & MFE-SSNet  \cite{huang2024mfe}           & MFE-SSNet  \cite{huang2024mfe}            & ST-GCN  & ST-GCN  & \multicolumn{2}{c|}{MFE-SSNet \cite{huang2024mfe}}              & 68.12          & 69.30          & 92.33          & 77.32 & 76.77                \\ 
 & YOLOP  \cite{wu2022yolop}           & YOLOP  \cite{wu2022yolop}             & ST-GCN  & ST-GCN  & \multicolumn{2}{c|}{YOLOP \cite{wu2022yolop}}       & 73.56        & 69.18         & 91.38          & 79.26 & 78.35
 \\  & PMANet \cite{liu2025umd}        & PMANet \cite{liu2025umd}            & 3DCNN    & 3DCNN    & \multicolumn{2}{c|}{PMANet \cite{liu2025umd}}     & 76.56      & 71.88   & 95.31  & 82.81  & 81.64
 \\ \midrule
\multirow{5}{*}{\begin{tabular}[c]{@{}c@{}}2D +\\ Sequential\end{tabular}} & Res18~\cite{he2016deep}+TransE     & Res34~\cite{he2016deep}+TransE~\cite{vaswani2017attention}     & MLP+TE  & MLP+TE  & \multicolumn{2}{c|}{PP-Res18+TransE~\cite{vaswani2017attention}}  & 70.83          & 67.32          & 90.54          & 79.97         & 77.17          \\
& Res18~\cite{he2016deep}+TransE~\cite{vaswani2017attention}     & Res34~\cite{he2016deep}+TransE~\cite{vaswani2017attention}     & MLP+TE  & MLP+TE  & \multicolumn{2}{c|}{Res34~\cite{he2016deep}+TransE}       & 72.65            & 67.08    & 86.63          & 78.46         & 76.21                  \\
 & Res34~\cite{he2016deep}+TransE~\cite{vaswani2017attention}     & Res50~\cite{he2016deep}+TransE~\cite{vaswani2017attention}     & MLP+TE  & MLP+TE  & \multicolumn{2}{c|}{Res50~\cite{he2016deep}+TransE~\cite{vaswani2017attention}}     & 70.24          & 65.65          & 82.57          & 77.29          & 73.94                 \\
 & VGG13~\cite{simonyan2014very}+TransE~\cite{vaswani2017attention}     & VGG16~\cite{simonyan2014very}+TransE~\cite{vaswani2017attention}     & MLP+TE  & MLP+TE  & \multicolumn{2}{c|}{VGG16~\cite{simonyan2014very}+TransE~\cite{vaswani2017attention}}     & 71.12          & 67.15         & 85.13          & 78.58         & 75.50                \\
 & VGG16~\cite{simonyan2014very}+TransE~\cite{vaswani2017attention}     & VGG19~\cite{simonyan2014very}+TransE~\cite{vaswani2017attention}     & MLP+TE  & MLP+TE  & \multicolumn{2}{c|}{VGG19~\cite{simonyan2014very}+TransE~\cite{vaswani2017attention}}     & 69.46          & 65.48         & 85.74          & 77.91         & 74.65                       \\ \midrule
\multirow{11}{*}{3D}  & MobileNet-V1~\cite{howard2017mobilenets}  & MobileNet-V1~\cite{howard2017mobilenets}  & ST-GCN   & ST-GCN   & \multicolumn{2}{c|}{MobileNet-V1~\cite{howard2017mobilenets}}  & 72.23          & 64.20          & 88.34          &77.83          & 75.65              \\
& MobileNet-V2~\cite{sandler2018mobilenetv2}  & MobileNet-V2~\cite{sandler2018mobilenetv2}  & ST-GCN   & ST-GCN   & \multicolumn{2}{c|}{MobileNet-V2~\cite{sandler2018mobilenetv2}}  & 68.47          & 61.74         & 86.54          & 78.66        & 73.85               \\
 & ShuffleNet-V1~\cite{zhang2018shufflenet} & ShuffleNet-V1~\cite{zhang2018shufflenet} & ST-GCN   & ST-GCN   & \multicolumn{2}{c|}{ShuffleNet-V1~\cite{zhang2018shufflenet}} & 72.41          & 68.97         & 90.64          &80.79         & 78.20              \\
 & ShuffleNet-V2~\cite{ma2018shufflenet} & ShuffleNet-V2~\cite{ma2018shufflenet} & ST-GCN   & ST-GCN   & \multicolumn{2}{c|}{ShuffleNet-V2~\cite{ma2018shufflenet}} & 70.94          & 64.04          & 89.33          & 78.98         & 75.82               \\
& 3D-Res18~\cite{hara2018can}         & 3D-Res34~\cite{hara2018can}         & ST-GCN   & ST-GCN   & \multicolumn{2}{c|}{3D-Res34~\cite{hara2018can}}         & 70.11          & 66.52          & 88.51          & 81.21          & 76.59                 \\
  & 3D-Res34~\cite{hara2018can}         & 3D-Res50~\cite{hara2018can}         & ST-GCN   & ST-GCN   & \multicolumn{2}{c|}{3D-Res50~\cite{hara2018can}}         & 69.13          & 63.05          & 87.82          & 79.31          & 74.83                \\
& C3D~\cite{tran2015learning}              & C3D~\cite{tran2015learning}              & ST-GCN   & ST-GCN   & \multicolumn{2}{c|}{C3D~\cite{tran2015learning}}              & 63.05          & 63.95         & 85.41          & 77.01        & 72.36                 \\
 & I3D~\cite{carreira2017quo}              & I3D~\cite{carreira2017quo}               & ST-GCN   & ST-GCN   & \multicolumn{2}{c|}{I3D~\cite{carreira2017quo} }              & 70.94          &66.17         & 87.68         & 79.81         & 76.15                \\
  & SlowFast~\cite{feichtenhofer2019slowfast}         & SlowFast~\cite{feichtenhofer2019slowfast}         & ST-GCN   & ST-GCN   & \multicolumn{2}{c|}{SlowFast~\cite{feichtenhofer2019slowfast}}         & 72.38          & 61.58         & 86.86          & 78.33          & 74.79       \\

 & TimeSFormer~\cite{bertasius2021space}      & TimeSFormer~\cite{bertasius2021space}      & ST-GCN   & ST-GCN   & \multicolumn{2}{c|}{TimeSFormer~\cite{bertasius2021space}}      & 74.87  & 65.18 & 92.12 & 78.81 & 77.75                 \\
   & Video Swin Transformer~\cite{liu2022video}         & Video Swin Transformer~\cite{liu2022video}        & 3DCNN    & 3DCNN    & \multicolumn{2}{c|}{Video Swin Transformer~\cite{liu2022video}}          & 73.44          & 65.63          & 93.75          & 75.00          & 76.96                \\ \midrule
 \rowcolor{gray!15}
\multirow{1}{*}{MMTL-UniAD} & MARNet         & MARNet             & 3DCNN    & 3DCNN    & \multicolumn{2}{c|}{MARNet}          & \underline{76.67}          & \underline{73.61}          & 93.91          & 85.00         & \underline{82.30}    \\
\rowcolor{gray!15}
\multirow{1}{*}{\textbf{UV-M$^3$TL}} & \textbf{MARNet}         & \textbf{MARNet}             & \textbf{3DCNN}    & \textbf{3DCNN}    & \multicolumn{2}{c|}{\textbf{MARNet}}          & \textbf{77.39}  & \textbf{73.82}          & \textbf{96.57}          & \textbf{87.07}          & \textbf{83.71}    \\
\bottomrule
\end{tabular}
}
\label{table1}
\vspace{-0.4cm}
\end{table*}

\subsection{Comparison with the State-of-the-Art}
\label{Comparison with the State-of-the-Art}

Table \ref{table1} presents the multi-task evaluation results of our proposed UV-M$^3$TL on the AIDE dataset, compared to the state-of-the-art methods categorized by their feature extraction dimensions. Following \cite{yang2023aide}, we group the comparison methods into three patterns: \textbf{2D} (using 2D models), \textbf{2D+Sequential} (combining 2D models with sequence models \cite{vaswani2017attention}), and \textbf{3D} (using 3D models, e.g., Video Swin Transformer \cite{liu2022video}, and 3D Implementations of MobileNet \cite{howard2017mobilenets} and ShuffleNet \cite{zhang2018shufflenet}). 

Our UV-M$^3$TL achieves the best performance across all four tasks, with an mAcc improvement of 1.41\%-13.50\% over other algorithms in these three patterns. We compare it with the baseline models released alongside the AIDE dataset (e.g., Res18~\cite{he2016deep}, VGG16~\cite{simonyan2014very}+TransE~\cite{vaswani2017attention}, MobileNet\cite{howard2017mobilenets,sandler2018mobilenetv2}, C3D~\cite{tran2015learning}) and their optimized versions with advanced backbones (e.g., CPVT~\cite{chu2023conditional}, CMT~\cite{Guo_2022_CVPR}, Video Swin Transformer~\cite{liu2022video}, HAT \cite{liu2024vision}), and recent representative MTL methods (e.g. YOLOP  \cite{wu2022yolop}, PMANet \cite{liu2025umd}, MFE-SSNet  \cite{huang2024mfe}, MTS-Mamba \cite{liu2025tem}). UV-M$^3$TL consistently outperforms all compared approaches, demonstrating its superiority in MTL.

In our unified framework using MARNet and 3DCNN as multimodal backbones for feature extraction, UV-M$^3$TL outperforms MMTL-UniAD \cite{liu2025mmtl} on all four tasks through integrating our developed DB-SCME and AFD-Loss. Notably, it achieves relative accuracy improvements of 2.66\% for traffic context recognition and 2.07\% for vehicle behavior recognition. These improvements validate that DB-SCME and AFD-Loss effectively alleviate negative transfer effects and enhance cross-task synergy. More detailed ablation results and analyses are provided in Section \ref{Ablation Studies on DB-SCME and AFD-Loss}.

\begin{table*}[t]
\caption{Comparisons of different methods on PASCAL-Context. * the reproduced results of work \cite{ye2023taskexpert} with a ViT-Large backbone.}
\label{tab:pascal_context_comp}
\centering
\renewcommand{\arraystretch}{1.1}
\resizebox{0.98\linewidth}{!}{
\begin{tabular}{c|cccccccc}
\Xhline{1px}
\multirow{2}{*}{\textbf{Pattern}} &
\multirow{2}{*}{\textbf{Method}} &
\multirow{2}{*}{\textbf{Publication}} &
\multirow{2}{*}{\textbf{Backbone}} &
\textbf{Semseg} & \textbf{Parsing} & \textbf{Saliency} &
\textbf{Normal} & \textbf{Boundary}\\
 & & & &
 mIoU$\uparrow$ & mIoU$\uparrow$ & maxF$\uparrow$ & mErr$\downarrow$ & odsF$\uparrow$ \\
\midrule
\multirow{9}{*}{CNN-based} & ASTMT \cite{maninis2019attentive} & CVPR'19 & -- & 68.00 & 61.10 &  65.70  &  14.70  &  72.40 \\
& PAD-Net \cite{xu2018pad} & CVPR'18 & HRNet18 & 53.60 & 59.60 & 65.80 & 15.30 & 72.50 \\
& PAD-Net* \cite{xu2018pad} & CVPR'18 & ViT-large & 78.01 & 67.12 & 79.21 & 14.37 & 72.60 \\
& MTI-Net \cite{vandenhende2020mti} & ECCV'20 & HRNet18 & 61.70 & 60.18 & 84.78 & 14.23 & 70.80 \\
& MTI-Net* \cite{vandenhende2020mti} & ECCV'20 & ViT-large & 78.31 & 67.40 & 84.75 & 14.67 & 73.00 \\
& ATRC \cite{bruggemann2021exploring} & ICCV'21 & HRNet18 & 62.69 & 59.42 & 84.70 & 14.20 & 70.96 \\
& ATRC-ASPP \cite{bruggemann2021exploring} & ICCV'21 & HRNet18 & 63.60 & 60.23 & 83.91 & 14.30 & 70.86 \\
& ATRC-BMTAS \cite{bruggemann2021exploring} & ICCV'21 & HRNet18 & 67.67 & 62.93 & 82.29 & 14.24 & 72.42 \\
& ATRC* \cite{bruggemann2021exploring} & ICCV'21 & ViT-large & 77.11 & 66.84 & 81.20 & 14.23 & 72.10 \\
\midrule
\multirow{7}{*}{Transformer-based} & InvPT \cite{ye2022inverted} & ECCV'22 & ViT-large & 79.03 & 67.61 & 84.81 & 14.15 & 73.00 \\
& InvPT++ \cite{ye2024invpt++}  & TPAMI'24 & ViT-large & 80.22 & 69.12 & 84.74 & 13.73 & 74.20 \\
& TaskPrompter \cite{ye2022taskprompter} & ICLR'23 & ViT-large & 80.89 & 68.89 & 84.83 & 13.72 & 73.50 \\
& TaskExpert \cite{ye2023taskexpert} & ICCV'23 & ViT-large & 80.64 & 69.42 & 84.87 & 13.56 & 73.30 \\
& MQTransformer \cite{xu2023multi} & TCSVT'23 & -- & 78.93 & 67.41 & 83.58 & 14.21 & 73.90 \\
& TSP-Transformer \cite{wang2024tsp} & WACV'24 & ViT-large & 81.48 & 70.64 & 84.86 & 13.69 & 74.80 \\
& MLoRE \cite{yang2024multi} & CVPR'24 & ViT-large & 81.41 & 70.52 & 84.90 & \underline{13.51} & 75.42 \\
\midrule
Diffusion-based & TaskDiffusion \cite{yangmulti} & ICLR'25 & ViT-large & 81.21 & 69.62 & 84.94 & 13.55 & 74.89 \\
\midrule
\multirow{2}{*}{Mamba-based} & MTMamba \cite{lin2024mtmamba} & ECCV'24 & Swin-Large & 81.11 & \underline{72.62} & 84.14 & 14.14 & \textbf{78.80} \\
& MTMamba++ \cite{lin2025mtmamba++} & TPAMI'25 & Swin-Large & \underline{81.94} & \textbf{72.87} & \underline{85.56} & 14.29 & \underline{78.60} \\
\midrule
\rowcolor{gray!15}
\textbf{Ours} & \textbf{UV-M$^3$TL}& -- & ViT-large & \textbf{82.23} & 71.18 & \textbf{85.78} & \textbf{13.44} & 77.06 \\
\bottomrule
\end{tabular}}
\end{table*}

\begin{table}[t]
\caption{Comparison results of different methods on NYUD-v2.}
\label{tab:nyud_comp}
\centering
\renewcommand{\arraystretch}{1.2}
\resizebox{1\linewidth}{!}{
\begin{tabular}{lccccc}
\Xhline{1px}
\multirow{2}{*}{\textbf{Method}} &
\multirow{2}{*}{\textbf{Backbone}} &
\multicolumn{1}{c}{\textbf{Semseg}} &
\multicolumn{1}{c}{\textbf{Depth}} &
\multicolumn{1}{c}{\textbf{Normal}} &
\multicolumn{1}{c}{\textbf{Boundary}} \\
 &  & mIoU$\uparrow$ & RMSE$\downarrow$ & mErr$\downarrow$ & odsF$\uparrow$ \\
\midrule
\multicolumn{6}{c}{\textit{CNN-based decoder}}\\
Cross-Stitch \cite{misra2016cross}     & --        & 36.34 & 0.6290 & 20.88 & 76.38 \\
PAP \cite{zhang2019pattern}            & ResNet    & 36.72 & 0.6178 & 20.82 & 76.42 \\
PSD \cite{zhou2020pattern}             & ResNet    & 36.69 & 0.6246 & 20.87 & 76.42 \\
PAD-Net~\cite{xu2018pad}               & HRNet18   & 36.61 & 0.6246 & 20.88 & 76.38 \\
MTI-Net~\cite{vandenhende2020mti}      & HRNet48   & 45.97 & 0.5365 & 20.27 & 77.86 \\
ATRC~\cite{bruggemann2021exploring}    & HRNet48   & 46.33 & 0.5363 & 20.18 & 77.94 \\ \midrule
\multicolumn{6}{c}{\textit{Transformer-based decoder}}\\
InvPT~\cite{ye2022inverted}            & ViT-large & 53.56 & 0.5183 & 19.04 & 78.10 \\
InvPT++~\cite{ye2022inverted}          & ViT-large & 53.85 & 0.5096 & 18.67 & 78.10 \\
TaskPrompter~\cite{ye2022taskprompter} & ViT-large & 55.30 & 0.5152 & 18.47 & 78.20 \\
TaskExpert~\cite{ye2023taskexpert}     & ViT-large & 55.35 & 0.5157 & 18.54 & 78.40 \\ 
MQTransformer \cite{xu2023multi}       & ViT-large & 54.84 & 0.5325 & 19.67 & 78.20 \\
TSP-Transformer \cite{wang2024tsp}     & ViT-large & 55.39 & 0.4961 & 18.44 & 77.50 \\
MLoRE \cite{yang2024multi}             & ViT-large & 55.96 & 0.5076 & 18.33 & 78.43 \\
\midrule
\multicolumn{6}{c}{\textit{Diffusion-based decoder}}\\
TaskDiffusion \cite{yangmulti}         & ViT-large & 55.65 & 0.5020 & 18.43 & 78.64 \\
\midrule
\multicolumn{6}{c}{\textit{Mamba-based decoder}}\\
MTMamba \cite{lin2024mtmamba}          & Swin-Large & 55.82 & 0.5066 & 18.63 & \underline{78.70} \\
MTMamba++ \cite{lin2025mtmamba++}      & Swin-Large & \underline{57.01} & \textbf{0.4818} & \underline{18.27} & \textbf{79.40} \\ \midrule
\rowcolor{gray!15}
\textbf{UV-M$^3$TL}                  & ViT-large  & \textbf{57.37} & \underline{0.4925} & \textbf{18.20} & 78.68 \\
\bottomrule
\end{tabular}}
\end{table}

\subsection{Evaluation on Other Multi-Task Benchmarks}
\label{Evaluation on Other Multi-Task Benchmarks}
To assess the versatility of UV-M$^3$TL, we conduct experiments on four fine perception benchmark datasets: NYUD-v2, PASCAL-Context, BDD100K, and CityScapes. These datasets represent the primary evaluation standards in panoptic driving perception and dense prediction. These experiments aim to examine the effectiveness of our proposed DB-SCME and AFD-Loss in MTL across different scenarios and task combinations.

Tables \ref{tab:pascal_context_comp} and \ref{tab:nyud_comp} present the experimental results of our method on NYUD-v2 and PASCAL-Context benchmarks. We compare our method with advanced MTL models on both datasets. Following \cite{lin2025mtmamba++}, methods are grouped into four categories based on their decoder architectures: CNN-based, Transformer-based, Diffusion-based, and Mamba-based approaches. Most existing dense prediction methods employ multi-task learning with pre-trained backbone encoders. Following \cite{chen2024multi}, we adopt ViT-Large as our encoder for fair comparison. Specifically, features extracted from its 6-th, 12-th, 18-th, and 24-th layers are fed into our designed DB-SCME module. The processed features are then directed to task-specific prediction heads to generate multi-task outputs. During training, we optimize the model using AFD-Loss as the supervision signal, with task-specific weights $\lambda_i$ set identical to those in \cite{chen2024multi}.

On the PASCAL-Context dataset (Table \ref{tab:pascal_context_comp}), our method establishes new state-of-the-art results in semantic segmentation, saliency detection, and surface normal estimation, while showing competitive performance on human parsing and object boundary detection. Across the five tasks (i.e., semantic segmentation, human parsing, saliency detection, surface normal estimation, and object boundary detection), our method consistently outperforms existing CNN-based, Transformer-based, and Diffusion-based approaches. Compared with MLoRE (one of the best performers among Transformer-based methods), our approach achieves improvements of +0.82 (mIoU$\uparrow$), +0.66 (mIoU$\uparrow$), +0.88 (maxF$\uparrow$), -0.07 (mErr$\downarrow$), and +1.64 (odsF$\uparrow$). The advantages are even more pronounced against Diffusion-based method (TaskDiffusion), with gains of +1.02 (mIoU$\uparrow$), +1.56 (mIoU$\uparrow$), +0.84 (maxF$\uparrow$), -0.11 (mErr$\downarrow$), and +2.17 (odsF$\uparrow$). Our approach further advances three task performances beyond the Mamba-based state-of-the-art (MTMamba++).

\begin{figure}[t]
    \centering
    \includegraphics[width=0.48\textwidth]{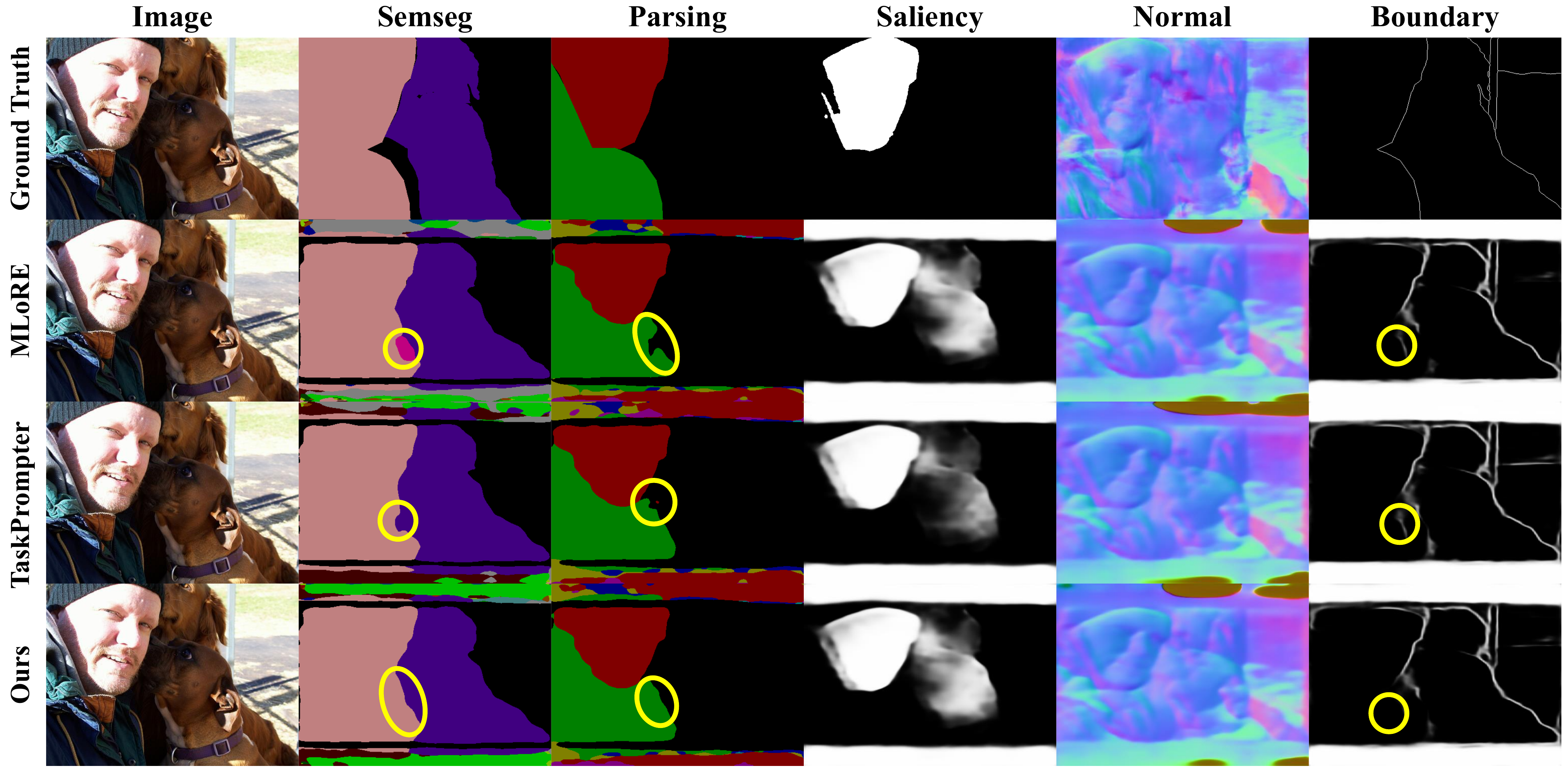}
    \vspace{-1em}
    \caption{Qualitative visualization results of different methods on the PASCAL-Context dataset, including MLoRE \cite{yang2024multi}, TaskPrompter \cite{ye2022taskprompter}, and ours. The regions highlighted by circles indicate areas where our method produces more accurate predictions, typically manifested in sharper object boundaries and more accurate recognition of overlapping target categories. Zoom in for more details.}
    \label{PASCAL}
\end{figure}

\begin{figure}[t]
    \centering
    \includegraphics[width=0.48\textwidth]{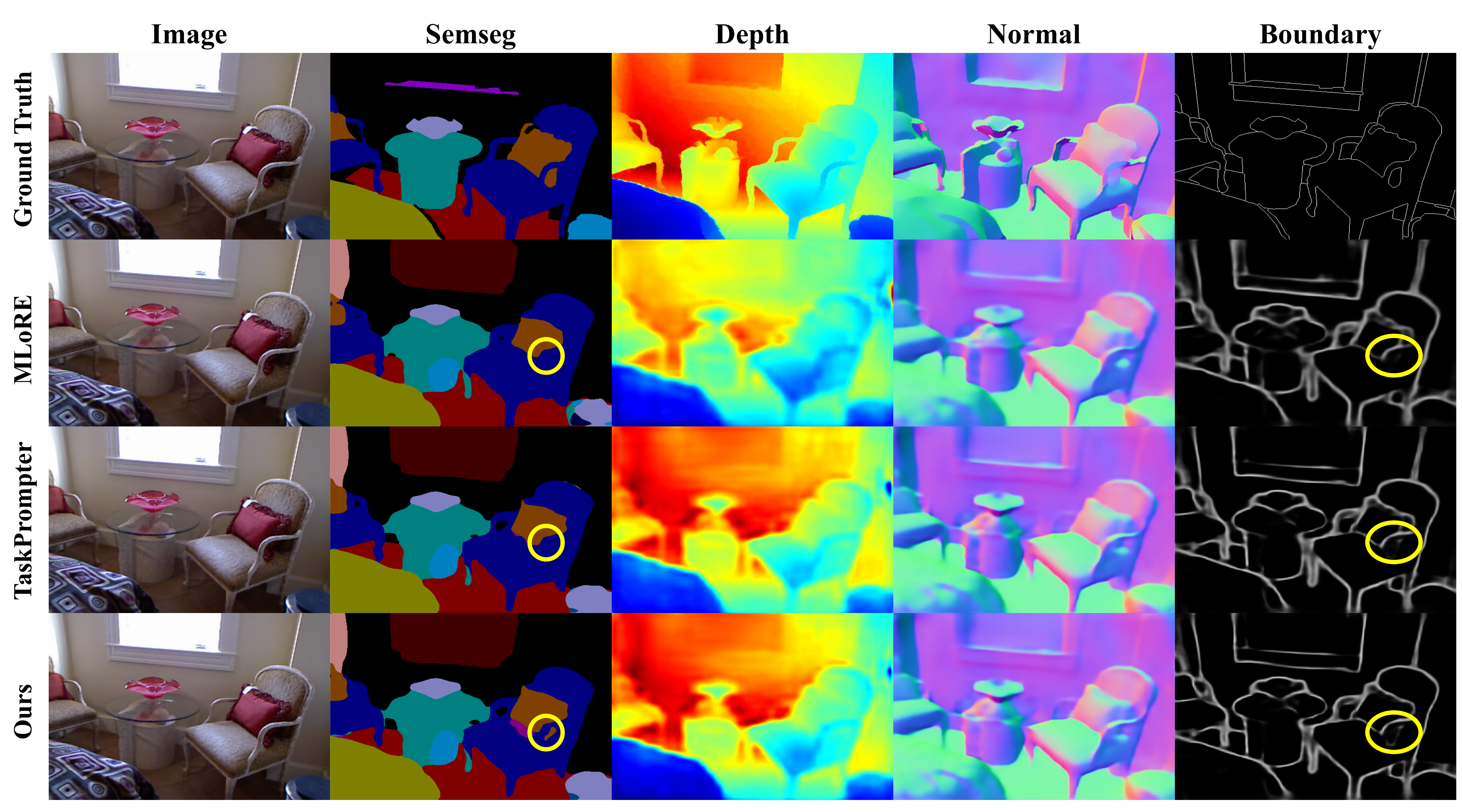}
    \vspace{-1em}
    \caption{Qualitative visualization results of different methods on the NYUD-v2 dataset, including MLoRE \cite{yang2024multi}, TaskPrompter \cite{ye2022taskprompter}, and ours. The regions highlighted by circles indicate areas with more accurate predictions. Zoom in for more details.}
    \label{NYUD}
\end{figure}

Table \ref{tab:nyud_comp} compares the performance on the NYUD-v2 dataset, with results consistent with those on PASCAL-Context. Our method outperforms all approaches on semantic segmentation and surface normal estimation, while remaining highly competitive in monocular depth estimation and object boundary detection. These results demonstrate the versatile and effectiveness of our method in multi-task dense prediction scenarios. We also show qualitative results of different methods on both datasets in Fig.\ref{PASCAL} and \ref{NYUD}. The proposed method achieves more accurate predictions across multiple tasks, with particularly notable improvements in challenging scenarios (e.g., object boundaries and overlapping targets).

Tables \ref{tab:combined_results} compare the performance of different methods on BDD100K. Specifically, we adopt key design elements from \cite{zhan2024yolopx}, including the task-specific prediction head architectures, base loss functions, and fixed weighting parameters $\lambda_i$. Building upon this foundation, we integrate our proposed DB-SCME and utilize AFD-Loss as the supervision signal for model training. The results show that incorporating DB-SCME and AFD-Loss leads to state-of-the-art performance in traffic object detection and drivable area segmentation, with competitive results in lane detection task.

\begin{table}[t]
\caption{Comparison of different methods on the BDD100K dataset.}
\label{tab:combined_results}
\centering
\small
\renewcommand{\arraystretch}{1.1}
\resizebox{1\linewidth}{!}{
\begin{tabular}{lcccccc}
\Xhline{1px}
\multirow{2}{*}{\textbf{Method}} &
\multicolumn{2}{c}{\textbf{Object detection}} &
\textbf{Semseg} &
\multicolumn{2}{c}{\textbf{Lane detection}} \\
& Recall$\uparrow$ & mAP50$\uparrow$ & mIoU$\uparrow$ & Accuracy$\uparrow$ & IoU$\uparrow$ \\
\midrule
Faster R-CNN~\cite{ren2015faster}  & 77.2 & 55.6 & --   & --    & --    \\
YOLOv5s~\cite{jocher2020yolov5}    & 86.8 & 77.2 & --   & --    & --    \\
MultiNet~\cite{teichmann2018multinet} & 81.3 & 60.2 & 71.6 & --    & --    \\
DLT-Net~\cite{qian2020dltnet}      & 89.4 & 68.4 & 71.3 & --    & --    \\
PSPNet~\cite{zhao2017pspnet}       & --   & --   & 89.6 & --    & --    \\
ENet~\cite{paszke2016enet}         & --   & --   & --   & 34.12 & 14.64 \\
SCNN~\cite{pan2018scnn}            & --   & --   & --   & 35.79 & 15.84 \\
ENet-SAD~\cite{hou2019enet_sad}    & --   & --   & --   & 36.56 & 16.02 \\
YOLOP \cite{wu2022yolop}        & 89.2 & 76.5 & 91.5 & 70.50 & 26.2  \\
HybridNets \cite{vu2022hybridnets}    & 92.8 & 77.3 & 90.5 & 85.4  & \textbf{31.6} \\
YOLOPv2 \cite{han2022yolopv2}      & 91.1 & 83.4 & \underline{93.2} & 87.3  & 27.2  \\
YOLOPX  \cite{zhan2024yolopx}      & \textbf{93.7} & 83.3 & \underline{93.2} & \textbf{88.6} & 27.2  \\
\rowcolor{gray!15}
UV-M$^3$TL  & \underline{93.5} & \textbf{83.9} & \textbf{93.4} & \underline{88.4} & \underline{28.4}  \\
\bottomrule
\end{tabular}}
\end{table}

Compared with the baseline model YOLOPX, our method consistently improves performance across all three tasks: a +0.6\% increase in the key metric mAP50 for traffic object detection, a +0.2\% (mIoU$\uparrow$) gain for drivable area segmentation, and a +1.2\% (IoU$\uparrow$) improvement for lane detection. While minor fluctuations occur in some metrics (e.g., recall and accuracy), the overall performance remains stable and surpassess the baseline. These results demonstrate the proposed method's strong performance in panoptic driving perception. By mitigating negative transfer and enhancing multi-task optimization, our designed DB-SCME and AFD-Loss deliver substantial gains to the MTL framework.

\begin{figure*}
\centering
\includegraphics[width=0.95\textwidth]{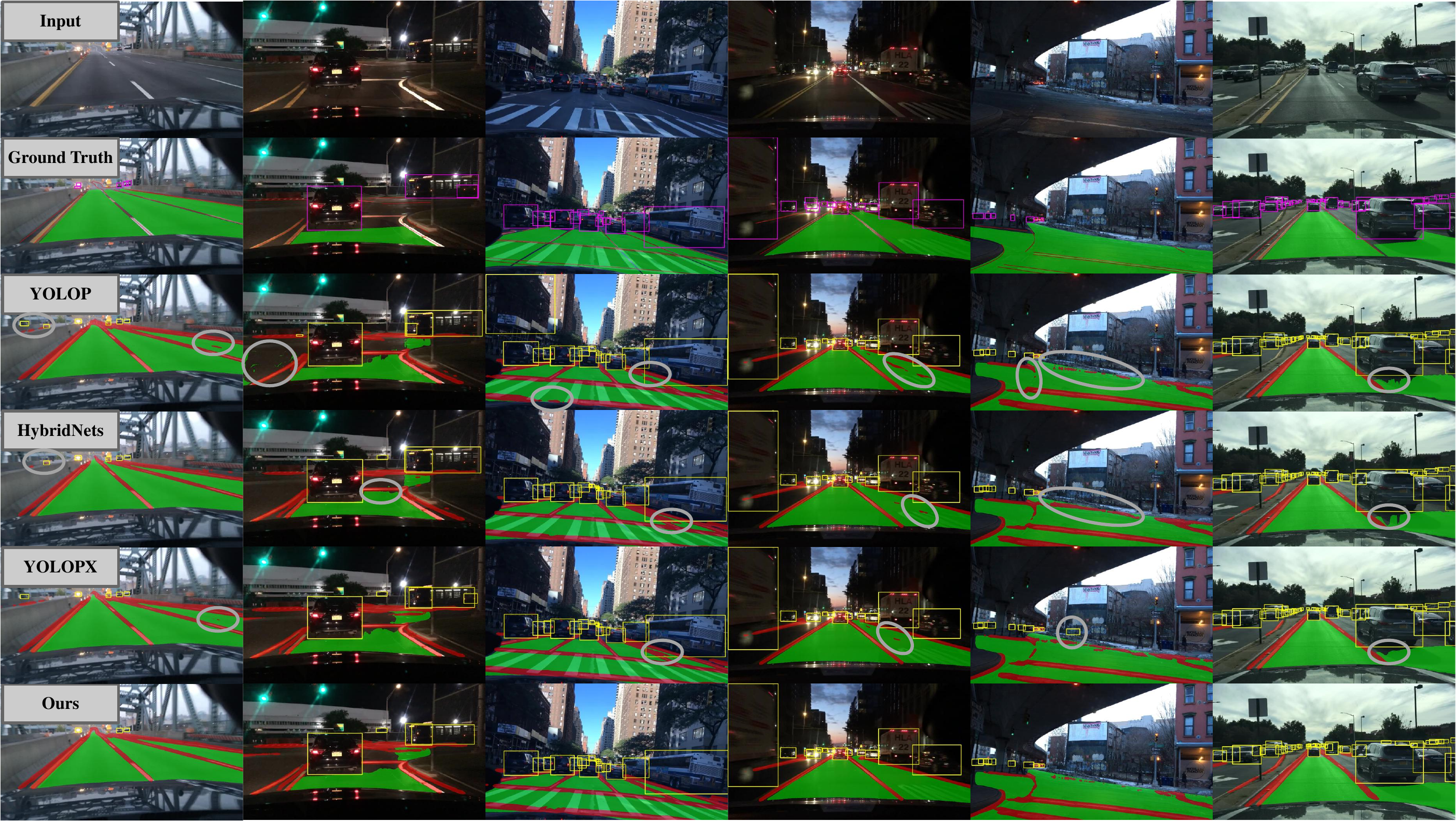}
\vspace{-1em}
\caption{Qualitative visualization results of different methods on the BDD100K dataset, including YOLOP \cite{wu2022yolop}, HybridNets \cite{vu2022hybridnets}, YOLOPX  \cite{zhan2024yolopx}, and ours. The regions highlighted by circles indicate areas where other methods produce incorrect predictions. Zoom in for more details.}
\vspace{-1em}
\label{BDD100K}
\end{figure*}

We present qualitative visualization comparisons of different methods on the BDD100K dataset in Fig. \ref{BDD100K}. We select diverse representative driving scenarios covering daytime/nighttime conditions and straight/curved roads. The results show that our method generates more accurate and stable predictions across various scene conditions, further validating its advantages for panoptic driving perception in multi-task scenarios.

\begin{table}[!t]
\centering
\setlength{\belowcaptionskip}{-0.10cm}
\caption{Comparisons of different methods on CityScapes. $^{\dagger}$ denotes that the model is trained using uncertainty weighting.}
\label{tab:cityscapes}
\renewcommand{\arraystretch}{1.1}
\resizebox{0.98\linewidth}{!}{
\begin{tabular}{l|cc|cc}
\Xhline{1px}
\multirow{2}{*}{Model} & \multicolumn{2}{c|}{Semseg} & \multicolumn{2}{c}{Depth} \\
\cline{2-5}
 & mIoU$\uparrow$ & PAcc$\uparrow$ & Abs.$\downarrow$ & Rel.$\downarrow$ \\
\midrule
Split-Wide~\cite{liu2019end} & 50.17 & 90.63 & 0.0167 & 44.73 \\
Split-Wide$^{\dagger}$ & 56.36 & 93.08 & 0.0130 & 28.72 \\
Split-Deep~\cite{liu2019end} & 49.85 & 88.69 & 0.0180 & 43.86 \\
Split-Deep$^{\dagger}$ & 54.08 & 91.26 & 0.0137 & 29.10 \\
Cross-Stitch~\cite{misra2016cross} & 50.08 & 90.33 & 0.0154 & 34.49 \\
Cross-Stitch$^{\dagger}$ & 56.29 & 93.08 & 0.0127 & 22.28 \\
MTAN~\cite{liu2019end} & 53.04 & 91.11 & 0.0144 & 33.63 \\
MTAN$^{\dagger}$ & 56.55 & 93.56 & 0.0140 & 24.64 \\
Dense~\cite{huang2017densely} & 51.91 & 90.89 & 0.0138 & 27.21 \\
Dense$^{\dagger}$ & 57.57 & 93.67 & \underline{0.0122} & 21.82 \\
\midrule
LSSA~\cite{sun2020learning} & 50.25 & 83.50 & 0.0162 & 35.96 \\
PCGrad~\cite{yu2020gradient} & 53.59 & 91.45 & 0.0171 & 31.34 \\
KD4MTL~\cite{li2020knowledge} & 52.71 & 91.54 & 0.0139 & 27.33 \\
$S^3$DMT-teacher~\cite{jha2021s} & 57.56 & 92.61 & 0.0124 & 23.51 \\
AdaMT-Net~\cite{jha2020adamt} & 61.91 & 94.01 & 0.0129 & 27.33 \\
XTasC-Net-ResNet34$^{\ddagger}$~\cite{nakano2021cross} & \underline{66.51} & 93.56 & \underline{0.0122} & \textbf{19.40} \\
MLwSGSU~\cite{lee2022multitask} & 53.80 & 90.96 & 0.0140 & 22.99 \\
MTAL~\cite{feng2022learning} & 54.10 & 91.61 & 0.0139 & 23.93 \\
UMT-Net \cite{chen2023umt} & 62.34 & \underline{95.18} & \textbf{0.0117} & 21.24 \\
\midrule
\rowcolor{gray!15}
UV-M$^3$TL & \textbf{68.59} & \textbf{96.17} & 0.0130 & \underline{19.82} \\
\Xhline{1px}
\end{tabular}}
\vspace{0.2em}
\end{table}

We evaluate our method with different advanced MTL models on the CityScapes dataset (Table \ref{tab:cityscapes}). Our method outperforms existing MTL baselines. For semantic segmentation, it achieves +2.08\% (mIoU$\uparrow$) and +0.99\% (PAcc$\uparrow$) improvements, while maintaining  competitive depth estimation accuracy.

\begin{table}[t]
\def\arraystretch{1.25}
\caption{Results of Multi-task Ablation Experiments for driver states and traffic context. "w/" indicates the use of the corresponding component or method, while "w/o" means the component or method was not used. Driver states include DER and DBR, while the traffic context includes TCR and VBR.}
\resizebox{\linewidth}{!}{%
\begin{tabular}{c|cccc|cccc}
\toprule
& \multicolumn{4}{c|}{\textbf{Task}} & \textbf{DER} & \textbf{DBR} & \textbf{TCR} & \textbf{VBR} \\ \cline{2-9}
\multirow{-2}{*}{\centering \textbf{Model}} & 
\multicolumn{2}{c}{\textbf{Driver States}} & 
\multicolumn{2}{c|}{\textbf{Traffic Context}} & 
\textbf{Acc} & \textbf{Acc} & \textbf{Acc} & \textbf{Acc} \\ 
\midrule

\multirow{3}{*}{\begin{tabular}[c]{@{}c@{}}MMTL- \\ UniAD\end{tabular}}
& \multicolumn{2}{c}{w/} &\multicolumn{2}{c|}{w/o} &  72.22 & 69.35 & - & - \\

& \multicolumn{2}{c}{w/o}& \multicolumn{2}{c|}{w/} & - & - & 90.41 & 80.63 \\

& \multicolumn{2}{c}{w/} & \multicolumn{2}{c|}{w/}  & \textbf{76.67} & \textbf{73.61} & \textbf{93.91} & \textbf{85.00} \\ \bottomrule

\multirow{3}{*}{UV-M$^3$TL}
& \multicolumn{2}{c}{w/} &\multicolumn{2}{c|}{w/o} &  73.41 & 69.29 & - & - \\

& \multicolumn{2}{c}{w/o}& \multicolumn{2}{c|}{w/} & - & - & 91.88 & 82.02 \\

& \multicolumn{2}{c}{w/} & \multicolumn{2}{c|}{w/}  & \textbf{77.39}  & \textbf{73.82}          & \textbf{96.57}          & \textbf{87.07} \\ \bottomrule

\end{tabular}
}
\vspace{-0.4cm}
\label{table15}
\end{table}

\subsection{Ablation Experiment}
\label{Ablation Experiment}
We perform ablation experiments on the AIDE dataset to evaluate how multi-task learning, multimodal data, and our key components (DB-SCME and AFD-Loss) contribute individually to model performance.

\subsubsection{Multi-task Learning}
To further analyze the synergy between driver state and traffic context tasks, and validate the necessity of their joint learning, we perform two ablation experiments using MMTL-UniAD and UV-M$^3$TL. The first experiment evaluates the impact of combining both task categories. Table \ref{table15} shows that excluding traffic context during training leads to 4.26\%–4.45\% accuracy drops in driver emotion/behavior recognition for MMTL-UniAD, and 3.98\%–4.53\% for UV-M$^3$TL. Conversely, training without driver state recognition reduces traffic context and vehicle behavior recognition accuracy by 3.50\%-4.37\% for MMTL-UniAD and by 4.69\%–5.05\% for UV-M$^3$TL. These results confirm bidirectional features benefit the two task categories and joint learning can improve model accuracy.

\begin{table}[t]
\centering
\caption{Results of Multi-task Ablation Experiments.}
\renewcommand{\arraystretch}{1.2} 
\resizebox{\linewidth}{!}{%
\begin{tabular}{>{\centering\arraybackslash}m{1.2cm}|>{\centering\arraybackslash}m{1.2cm}|>{\centering\arraybackslash}m{4cm}|>{\centering\arraybackslash}m{0.6cm} >{\centering\arraybackslash}m{0.6cm} >{\centering\arraybackslash}m{0.6cm} >{\centering\arraybackslash}m{0.6cm}}
\toprule
& & & \textbf{DER} & \textbf{DBR} & \textbf{TCR} & \textbf{VBR} \\ \cline{4-7}
\multirow{-2}{*}{\centering \textbf{Method}} & \multirow{-2}{*}{\centering \textbf{Config}} & \multirow{-2}{*}{\centering \textbf{Task Selection}} & \textbf{Acc} & \textbf{Acc} & \textbf{Acc} & \textbf{Acc} \\ 
\midrule

\multirow{4}{*}{STL}
& w/o & DBR \ \& \ TCR \ \& \ VBR & 70.56 & - & - & - \\

& w/o & DER \ \& \ TCR \ \& \ VBR & - & 68.12 & - & - \\

& w/o & DER \ \& \ DBR  \ \& \ VBR & - & - & 89.93 & - \\

& w/o & DER \ \& \ DBR \ \& \ TCR & - & - & - & 78.87 \\ \midrule
\multirow{5}{*}{\begin{tabular}[c]{@{}c@{}}MMTL- \\ UniAD\end{tabular}}
& w/ & DBR \ \& \ TCR \ \& \ VBR & - & 64.19 & 92.36 & 83.71 \\

& w/ & DER \ \& \ TCR \ \& \ VBR & 73.01 & - & 91.73 & 84.56 \\ 

& w/ & DER \ \& \ DBR \ \& \ VBR & 75.32 & 70.83 & - & 79.85 \\

& w/ & DER \ \& \ DBR \ \& \ TCR & 74.62 & 67.24 & 89.03 & - \\
& w/ & DER \ \& \ DBR \ \& \ TCR \ \& \ VBR
& \textbf{76.67} & \textbf{73.61} & \textbf{93.91} & \textbf{85.00} \\ \midrule
\multirow{5}{*}{UV-M$^3$TL}
& w/ & DBR \ \& \ TCR \ \& \ VBR & - & 67.88 & 95.15 & 83.90 \\

& w/ & DER \ \& \ TCR \ \& \ VBR & 72.35 & - & 94.72 & 83.74 \\ 

& w/ & DER \ \& \ DBR \ \& \ VBR & 74.88 & 69.64 & - & 80.36 \\

& w/ & DER \ \& \ DBR \ \& \ TCR & 75.79 & 68.31 & 91.65 & - \\
& w/ & DER \ \& \ DBR \ \& \ TCR \ \& \ VBR
& \textbf{77.39}  & \textbf{73.82}          & \textbf{96.57}          & \textbf{87.07} \\ \midrule

\end{tabular}
}
\label{table3}
\vspace{-0.5cm}
\end{table}

The second ablation experiment examines task interactions through two configurations: single-task training and exclusion of one task (while retaining the other three). Since the dedicated MTL modules proposed in this work are inapplicable to single-task scenarios, we adopt a unified framework where the multimodal features extracted by MARNet and 3DCNN are concatenated and then fed into task-specific prediction heads. As shown in Table \ref{table3}, UV-M$^3$TL exhibits performance decline of 5.7\%-8.2\% across all tasks under single-task training. Moreover, excluding any one of the four tasks degrades the accuracy of remaining tasks. These results highlight both the benefits of joint learning over four tasks and the inherent interconnections among tasks.

\begin{table}[t]
\setlength{\tabcolsep}{7pt}
\centering
\caption{Ablation experiment results of DB-SCME and AFD-Loss.}
\resizebox{\linewidth}{!}{%
\begin{tabular}{c|cc|cccc|c}
\toprule                                              
& & & \textbf{DER} & \textbf{DBR} & \textbf{TCR} & \textbf{VBR} &  \\ \cline{4-7}
\multirow{-2}{*}{\centering \textbf{Method}} & \multirow{-2}{*}{\centering \textbf{DB-SCME}} & \multirow{-2}{*}{\centering \textbf{AFD-Loss}} & \centering \textbf{Acc} & \centering \textbf{Acc} & \centering \textbf{Acc} & \centering \textbf{Acc} & \multirow{-2}{*}{\centering \textbf{mAcc}}\\ 
\midrule

\multirow{3}{*}{Contrast} & w/o  &  w/o  &  68.99     & 67.14     & 88.43      & 79.57      &  76.03    \\
& w/   &  w/o  & 77.11      & 73.64     & 95.67      & 86.42      &  83.21    \\
& w/o  &  w/   & 71.49     & 70.21      & 91.86      & 82.53   & 79.02 \\ \midrule
\multirow{1}{*}{\textbf{Ours}} &
w/   &  w/  & \textbf{77.39}  & \textbf{73.82}          & \textbf{96.57}          & \textbf{87.07}          & \textbf{83.71}     \\

\bottomrule
\end{tabular}
}
\label{table4}
\end{table}

\subsubsection{DB-SCME and AFD-Loss}
\label{Ablation Studies on DB-SCME and AFD-Loss}

Within UV-M$^3$TL, we conducted ablation experiments to evaluate the contributions of DB-SCME and AFD-Loss. Specifically, we replaced DB-SCME with a simple concatenation operation and AFD-Loss with a direct summation of the losses from each task. As shown in Table \ref{table4}, removing DB-SCME or AFD-Loss results in a significant performance drop of 0.5\%-7.68\% in mAcc, with accuracy declining consistently across all four tasks. The performance degradation primarily stems from the alternative components' inability to effectively model key shared and task-specific features, thus failing to capture both commonalities and discrepancies across tasks. Moreover, they overlook shifts in task learning dynamics during training. These limitations not only hinder cross-task synergy but also amplify negative transfer by increasing task conflicts.

\begin{table}[t]
\setlength{\tabcolsep}{7pt}
\centering
\caption{Ablation experiment results of spatial self-attention and channel self-attention in DB-SCME.}
\resizebox{\linewidth}{!}{%
\begin{tabular}{c|cc|cccc|c}
\toprule                                              
& & & \textbf{DER} & \textbf{DBR} & \textbf{TCR} & \textbf{VBR} &  \\ \cline{4-7}
\multirow{-2}{*}{\centering \textbf{Method}} & \multirow{-2}{*}{\centering \textbf{Spatial}} & \multirow{-2}{*}{\centering \textbf{Channel}} & \centering \textbf{Acc} & \centering \textbf{Acc} & \centering \textbf{Acc} & \centering \textbf{Acc} & \multirow{-2}{*}{\centering \textbf{mAcc}}\\ 
\midrule

\multirow{2}{*}{Contrast}
& w/   &  w/o  & 75.54      & 72.19     & 94.95      & 85.77      &  82.11    \\
& w/o  &  w/   & 76.88     & 73.69      & 95.10      & 86.02   & 82.92 \\ \midrule
\multirow{1}{*}{\textbf{Ours}} &
w/   &  w/  & \textbf{77.39}  & \textbf{73.82}          & \textbf{96.57}          & \textbf{87.07}          & \textbf{83.71}     \\

\bottomrule
\end{tabular}
}
\label{DB}
\end{table}

\begin{table}[t]
\setlength{\tabcolsep}{7pt}
\centering
\caption{Ablation experiment results of $\mathcal{L}_{\mathrm{D\_task}}$ and $\mathcal{L}_{\mathrm{decouple}}$.}
\resizebox{\linewidth}{!}{%
\begin{tabular}{c|cc|cccc|c}
\toprule                                              
& & & \textbf{DER} & \textbf{DBR} & \textbf{TCR} & \textbf{VBR} &  \\ \cline{4-7}
\multirow{-2}{*}{\centering \textbf{Method}} & \multirow{-2}{*}{\centering \textbf{$\mathcal{L}_{\mathrm{D\_task}}$}} & \multirow{-2}{*}{\centering \textbf{$\mathcal{L}_{\mathrm{decouple}}$}} & \centering \textbf{Acc} & \centering \textbf{Acc} & \centering \textbf{Acc} & \centering \textbf{Acc} & \multirow{-2}{*}{\centering \textbf{mAcc}}\\ 
\midrule

\multirow{3}{*}{Contrast}
& w/o   &  w/o & 77.11      & 73.64     & 95.67      & 86.42      &  83.21 \\
& w/   &  w/o  & 77.30      & 73.71     & 96.09      & 87.00      &  83.53    \\
& w/o  &  w/   & 77.18     & 73.48      & 95.89      & 86.61   & 83.29 \\ \midrule
 \multirow{1}{*}{\textbf{Ours}} &
w/   &  w/  & \textbf{77.39}  & \textbf{73.82}          & \textbf{96.57}          & \textbf{87.07}          & \textbf{83.71}     \\

\bottomrule
\end{tabular}
}
\label{AFDL}
\end{table}

\begin{figure}[t]
    \centering
    \includegraphics[width=0.48\textwidth]{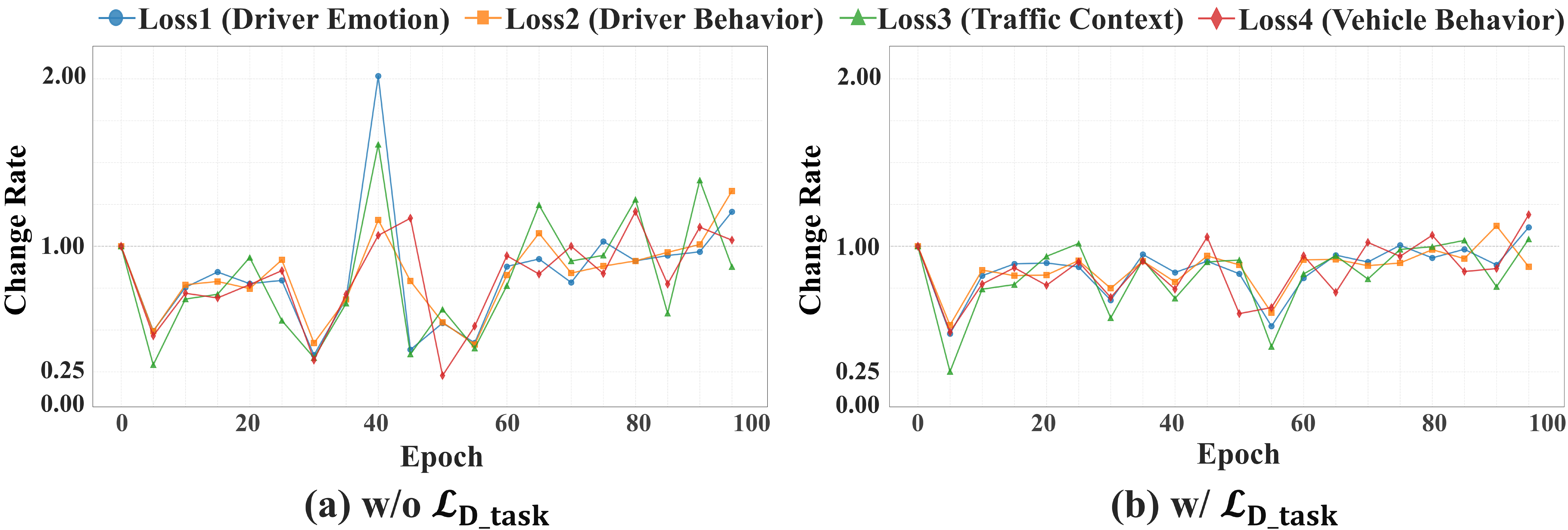}
    \vspace{-1em}
    \caption{Loss curves during joint training. (a) and (b) show the loss variations of four tasks without and with $\mathcal{L}_{\mathrm{D\_task}}$, respectively. Zoom in for more details.}
    \label{loss}
\end{figure}

\begin{table}[t]
\setlength{\tabcolsep}{7pt}
\renewcommand{\arraystretch}{1.2}
\centering
\caption{Results of Multimodal Ablation  Experiments for Four Tasks. G+P represents gesture and posture joint data.}
\resizebox{\linewidth}{!}{%
\begin{tabular}{c|cccl|cccc|c}
\toprule
& \multicolumn{4}{c|}{\textbf{Multimodal Data}} & \textbf{DER} & \textbf{DBR} & \textbf{TCR} & \textbf{VBR} &  \\ \cline{2-9}
\multirow{-2}{*}{\centering \textbf{Method}} & \multicolumn{2}{c}{\centering \textbf{Face+Body}} & \centering \textbf{G+P} & \textbf{Scene} & \textbf{Acc} & \textbf{Acc} & \textbf{Acc} & \textbf{Acc} & \multirow{-2}{*}{\centering \textbf{mAcc}}\\ 
\midrule

\multirow{3}{*}{Contrast}

&\multicolumn{2}{c}{w/o}  &  w/o    &\multicolumn{1}{c|}{w/} & 69.42   & 64.68    & 94.03    & 83.54   &  77.92   \\

&\multicolumn{2}{c}{w/o}  &\multicolumn{1}{c}{w/}        &\multicolumn{1}{c|}{w/o}     & 70.13          & 60.22          & 65.79          & 45.38  &  60.38       \\

&\multicolumn{2}{c}{w/}  &    w/o    &\multicolumn{1}{c|}{w/o}     & 74.55          & 72.29          & 89.74         & 81.33  &    79.48     \\

\midrule 
\multirow{1}{*}{\textbf{Ours}}                                   & \multicolumn{2}{c}{w/}           & \multicolumn{1}{c}{w/}  
& \multicolumn{1}{c|}{w/}          & \textbf{77.39}  & \textbf{73.82}          & \textbf{96.57}          & \textbf{87.07}          & \textbf{83.71}                            \\ 
 \bottomrule
\end{tabular}
}
\label{table2}
\vspace{-0.4cm}
\end{table}

The experimental results show that DB-SCME and AFD-Loss synergistically contribute to optimal performance. Specifically, they enhance the extraction of both task-shared and task-specific features, improving cross-task collaboration. Moreover, by modeling the dynamic evolution of task learning during training, they stabilize the multi-task optimization process, resulting in significant performance and broad applicability gains for MTL.

To further analyze the performance gains from DB-SCME and AFD-Loss, we conduct ablation studies on their key components. Table \ref{DB} reports the ablation results for the two core components of DB-SCME: spatial self-attention and channel self-attention. The results show that focusing only on spatial positional information or semantic channel information is insufficient to obtain the best feature representations. Only by modeling both types jointly can the model achieve more effective feature learning, thus improving the MTL model's overall performance.

Table \ref{AFDL} presents the ablation results of $\mathcal{L}_{\mathrm{D\_task}}$ and $\mathcal{L}_{\mathrm{decouple}}$ in AFD-Loss. Experiments show that using $\mathcal{L}_{\mathrm{D\_task}}$ or $\mathcal{L}_{\mathrm{decouple}}$ alone yields stable performance gains, demonstrating the effectiveness of leveraging task learning dynamics and reducing redundancy between task-shared and task-specific features. Combining the two components achieves the best performance, highlighting the benefits of their synergistic interaction. 

Furthermore, Fig. \ref{loss} (a) and (b) illustrate the loss change rate curves of four tasks during joint training without and with the incorporation of $\mathcal{L}_{\mathrm{D\_task}}$, respectively. The loss change rate is defined as the ratio between the loss value at the next epoch and that at the current epoch. A smaller value indicates a faster learning rate of the corresponding task at the current training stage, thereby providing a more intuitive reflection of the learning dynamics of different tasks during training. It can be observed that, in the absence of $\mathcal{L}_{\mathrm{D\_task}}$, the loss change rates of different tasks exhibit substantial fluctuations throughout the training process. In contrast, introducing $\mathcal{L}_{\mathrm{D\_task}}$ leads to a significantly more stable training behavior, with the loss change rates rarely exceeding 1, indicating that undesirable loss growth is effectively suppressed. These results further validate the effectiveness of $\mathcal{L}_{\mathrm{D\_task}}$ in enhancing the learning efficiency and optimization stability of MTL.

\subsubsection{Multimodal Data}
\label{Ablation studies on multimodal data}
To assess individual modality contribution, we performed ablation experiments with three distinct input configurations: (i) driver's facial and body sequential images, (ii) gesture and body joint data, and (iii) multi-view sequential images (i.e., right, left, front, inside views). We train the model separately on each configuration, with comparative results shown in Table \ref{table2}. The results show decreased mAcc and accuracy in all tasks when only one modality was used, underscoring the value of multimodal data for multi-task recognition. This demonstrates the multimodal data's critical role in improving multi-task learning performance, as well as the need for MTL architectures designed to integrate multimodal inputs.

\section{Conclusions}

This paper presents UV-M$^3$TL, a unified and versatile multi-modal multi-task learning framework for joint recognition of driver emotions/behaviors, traffic context, and vehicle behaviors. The framework integrates two key innovations: DB-SCME for extracting task-shared and task-specific features and AFD-Loss to stabilize joint optimization across heterogeneous tasks, thereby enhancing cross-task synergy while mitigating negative transfer. UV-M$^3$TL not only achieves state-of-the-art performance across all four tasks on the coarse perception AIDE dataset, but also demonstrates excellent performance on fine perception benchmarks (BDD100K, CityScapes, NYUD-v2, PASCAL-Context), validating its effectiveness and versatility. Ablation studies further reveal that jointly learning driver states and traffic context tasks facilitates mutual feature sharing, leading to significant accuracy improvements. We expect UV-M$^3$TL and its core components to serve as a robust baseline for future research in ADAS-oriented multimodal multi-task learning, contributing to more adaptable systems.

\bibliographystyle{IEEEtran}
\bibliography{TotalBib}

\begin{IEEEbiography}[{\includegraphics[width=1in,height=1.25in,clip,keepaspectratio]{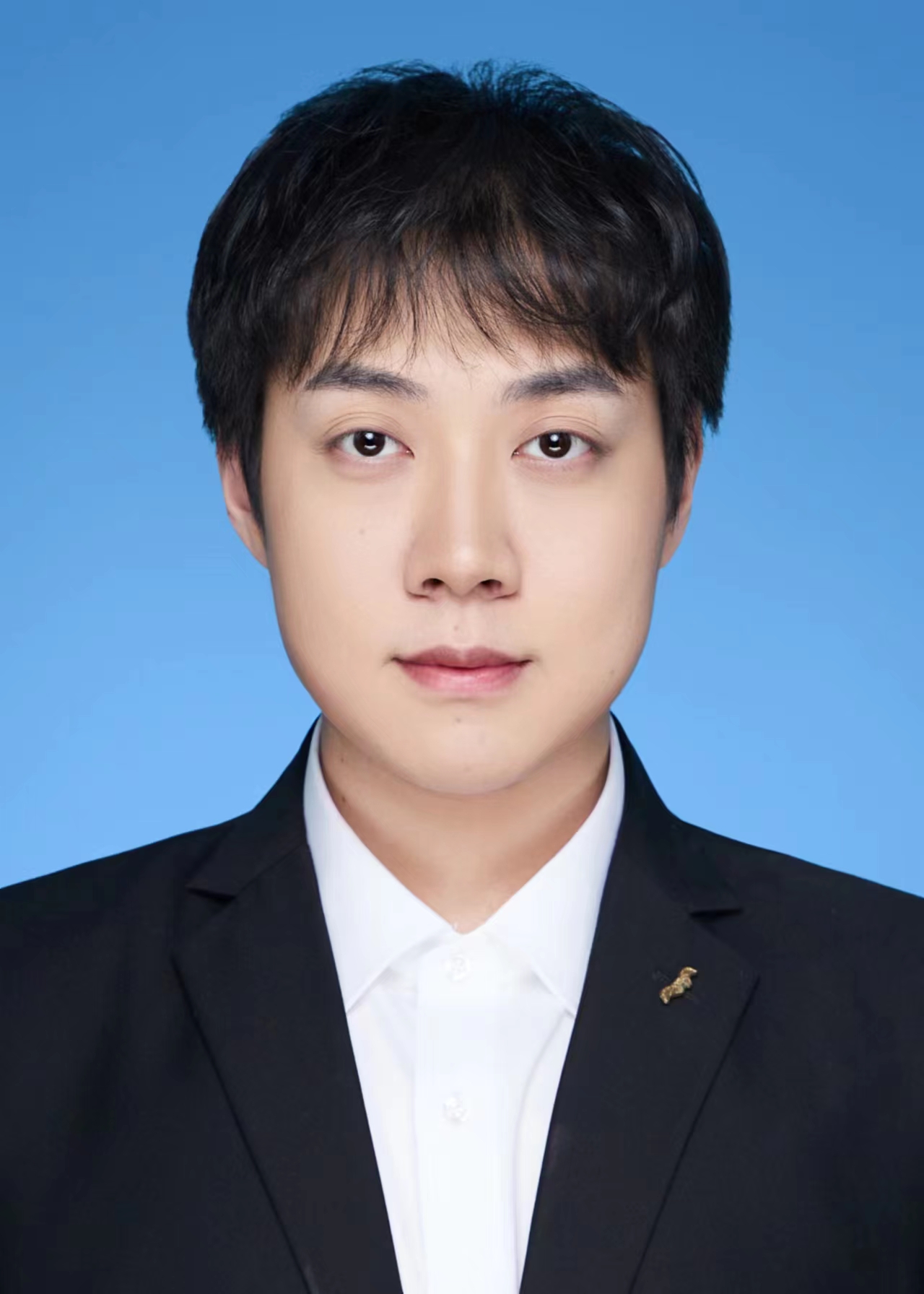}}]{Wenzhuo Liu (Graduate Student Member, IEEE)}
was born in Jinan, Shandong Province, China in 1999. He received his master's degree in computer science and technology from China University of Mining and Technology (Beijing). He was a joint student at the State Key Laboratory of Intelligent Green Vehicles and Mobility, School of Vehicle and Mobility, Tsinghua University for three years. Currently, he is a PhD candidate at the State Key Laboratory of Intelligent Unmanned Systems Technology, Beijing Institute of Technology (BIT), China. His research interests include multi-task learning, autonomous driving perception, multi-modal fusion, and driver intention recognition.
\end{IEEEbiography}

\begin{IEEEbiography}[{\includegraphics[width=1in,height=1.25in,clip,keepaspectratio]{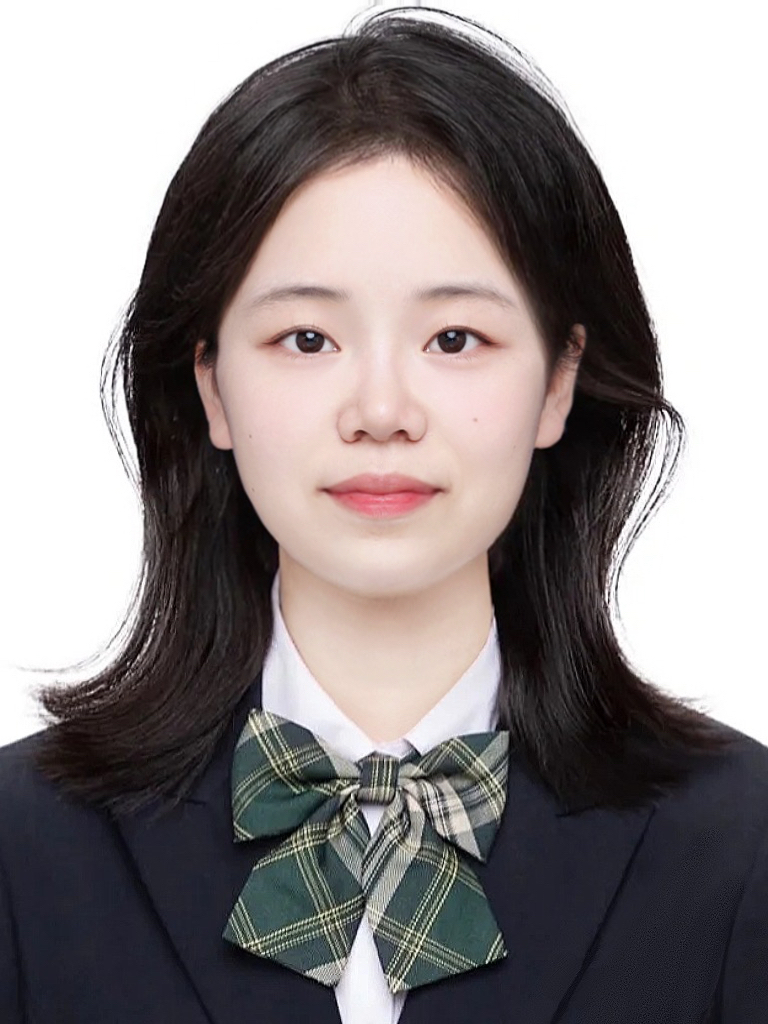}}]{Qiannan Guo}
was born in Jiaozuo, Henan Province, China. She received her Master’s degree from the College of Information and Electrical Engineering, China Agricultural University. She worked as a research assistant at the State Key Laboratory of Intelligent Technology and Systems, Department of Computer Science and Technology, Tsinghua University. She is currently a Ph.D. candidate in the School of Artificial Intelligence at Beijing Normal University. Her research focuses on computer vision, with primary interests in human behavior understanding, autonomous driving, and multi-task learning.
\end{IEEEbiography}

\begin{IEEEbiography}[{\includegraphics[width=1in,height=1.25in,clip,keepaspectratio]{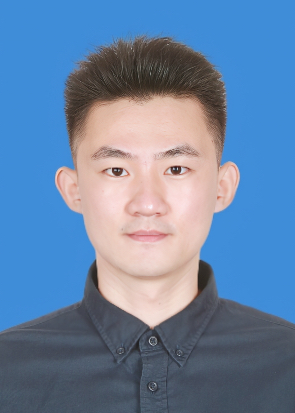}}]{Zhen Wang}
was born in Zaozhuang, Shandong Province, China in 1995. He received his master’s degree in mechanical engineering from Beijing Information Science and Technology University. He is currently pursuing his Ph.D. in the Energy and Transportation Domain at Beijing Institute of Technology (BIT), China. His research interests include multi-modal fusion and optical fiber sensing for electric vehicles.
\end{IEEEbiography}

\begin{IEEEbiography}[{\includegraphics[width=1in,height=1.25in,clip,keepaspectratio]{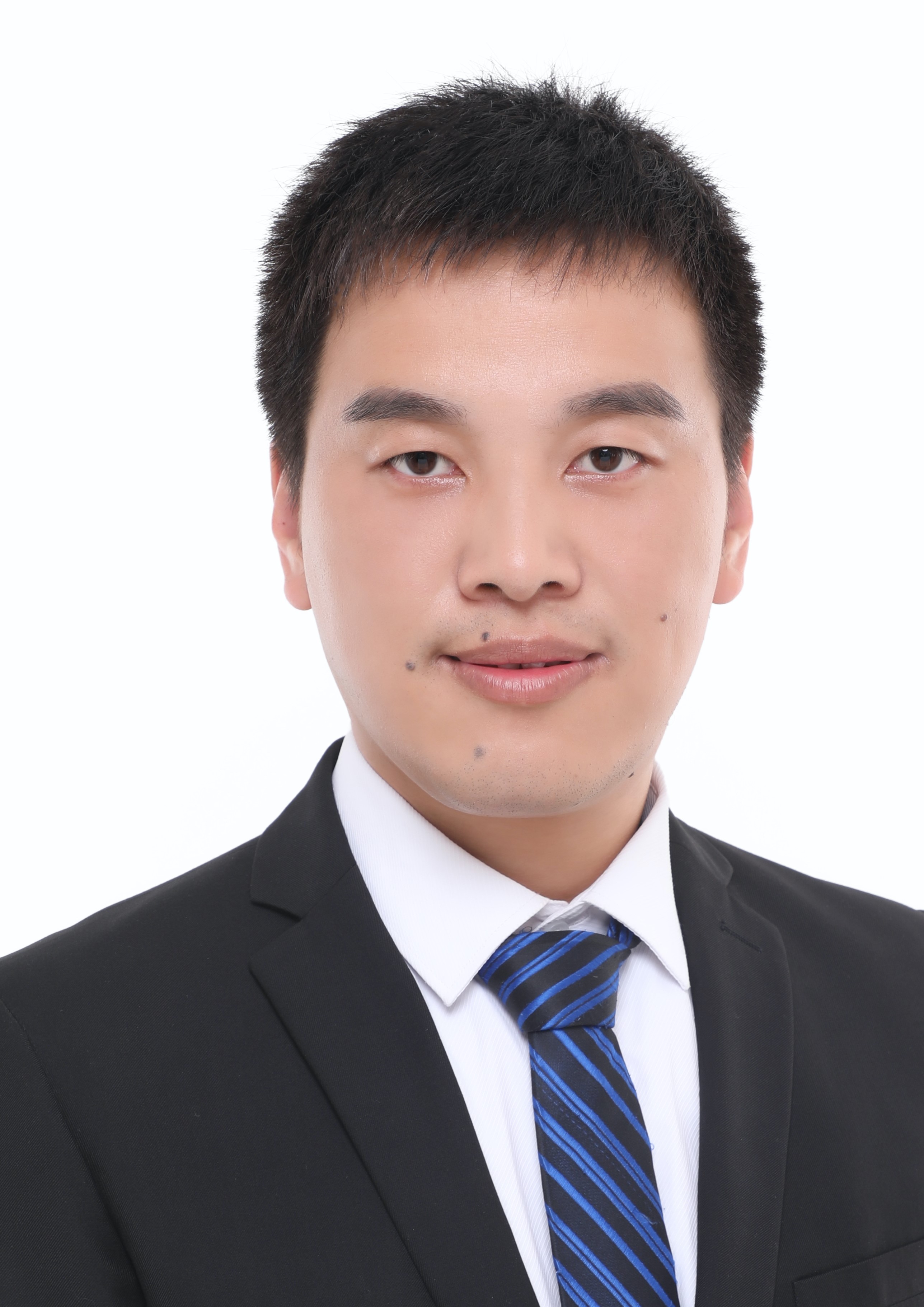}}]{Wenshuo Wang (Member, IEEE)}
 (SM'15-M'18) received his Ph.D. degree in mechanical engineering from the Beijing Institute of Technology (BIT) in 2018.  Presently, he is a Full Professor at the School of Mechanical Engineering, BIT, Beijing, China. Before his role at BIT, he completed Postdoctoral fellowships at McGill University, Carnegie Mellon University (CMU), and UC Berkeley between 2018 and 2023. Furthermore, from 2015 to 2018, he served as a Research Assistant at UC Berkeley and the University of Michigan, Ann Arbor. His research interests focus on Bayesian nonparametric learning, human driver models, human–vehicle interaction, ADAS, and autonomous vehicles.
\end{IEEEbiography}

\begin{IEEEbiography}[{\includegraphics[width=1in,height=1.25in,clip,keepaspectratio]{{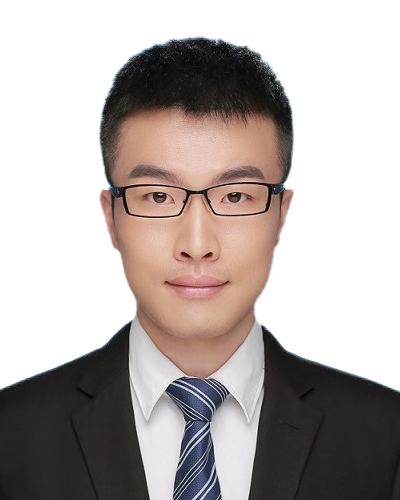}}}]{Lei Yang (Member, IEEE)} received his M.S. degree from the Robotics Institute at Beihang University, in 2018. and the Ph.D. degree from the School of Vehicle and Mobility, Tsinghua University, in 2024. From 2018 to 2020, he joined the Autonomous Driving R\&D Department of JD.COM as an algorithm researcher. Currently, he is a research fellow with the School of Mechanical and Aerospace Engineering, Nanyang Technological University, Singapore. His current research interests include autonomous driving, 3D scene understanding and world model.
\end{IEEEbiography}

\begin{IEEEbiography}[{\includegraphics[width=1in,height=1.25in,clip,keepaspectratio]{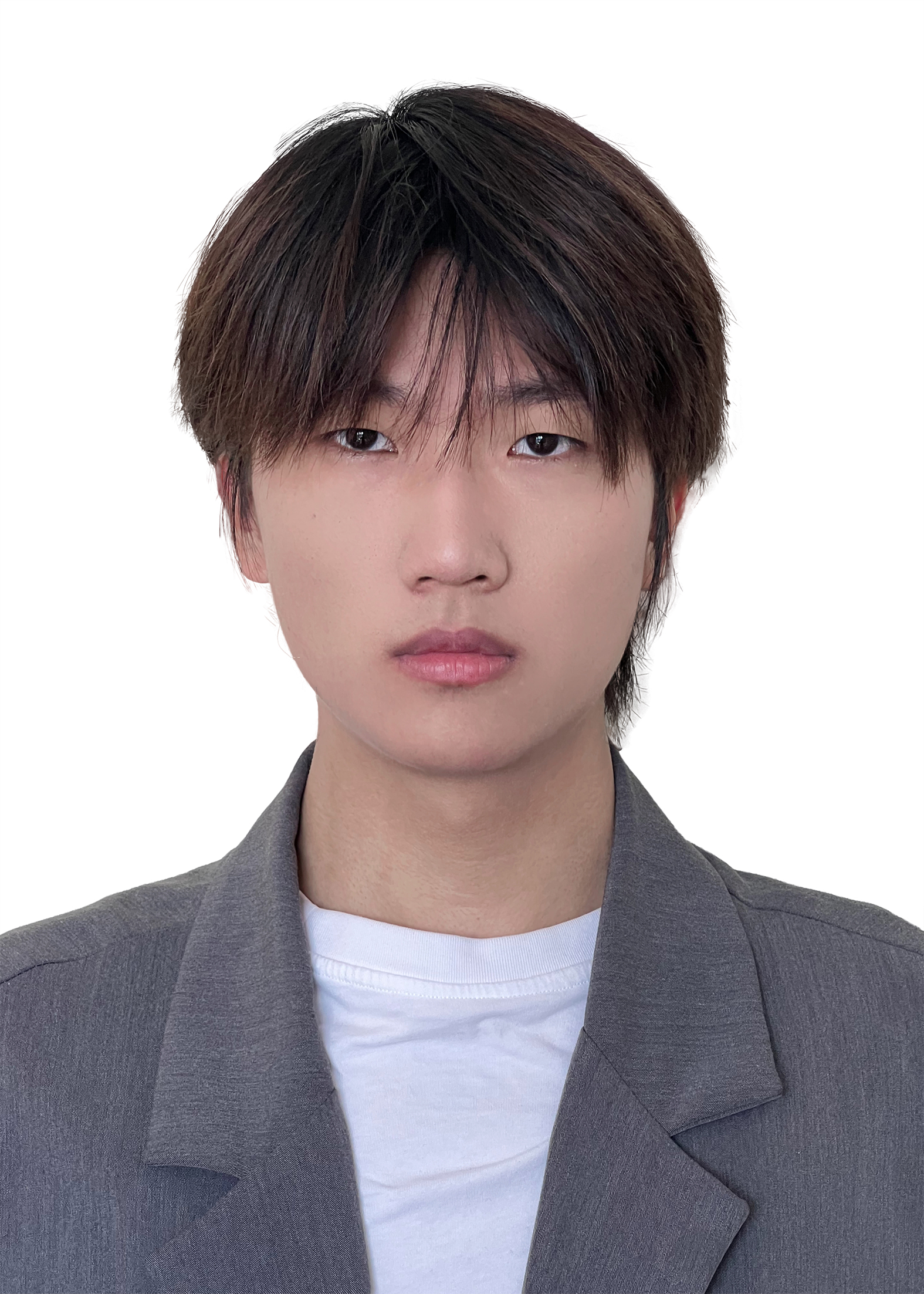}}]{Yicheng Qiao}
was born in Kaifeng, Henan Province, China, in 2001. He will begin his Ph.D. studies in Computer Science at Michigan State University in 2026. He was a joint undergraduate student at the State Key Laboratory of Intelligent Technology and Systems, Department of Computer Science and Technology, Tsinghua University. His research interests include large language models, multimodal learning, and agentic learning.
\end{IEEEbiography}

\begin{IEEEbiography}[{\includegraphics[width=1in,height=1.25in,clip,keepaspectratio]{{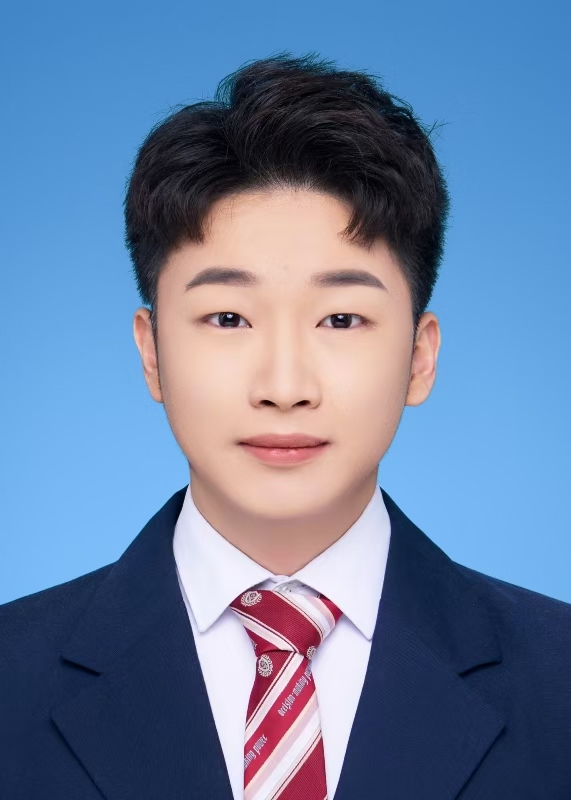}}}]{Lening Wang} 
is currently pursuing the M.S. degree at the School of Transportation Science and Engineering and the State Key Lab of Intelligent Transportation System, Beihang University, Beijing, China. His current research interests include Embodied Transportation System (ETS), autonomous driving, and world model.
\end{IEEEbiography}

\begin{IEEEbiography}[{\includegraphics[width=1in,height=1.25in,clip,keepaspectratio]{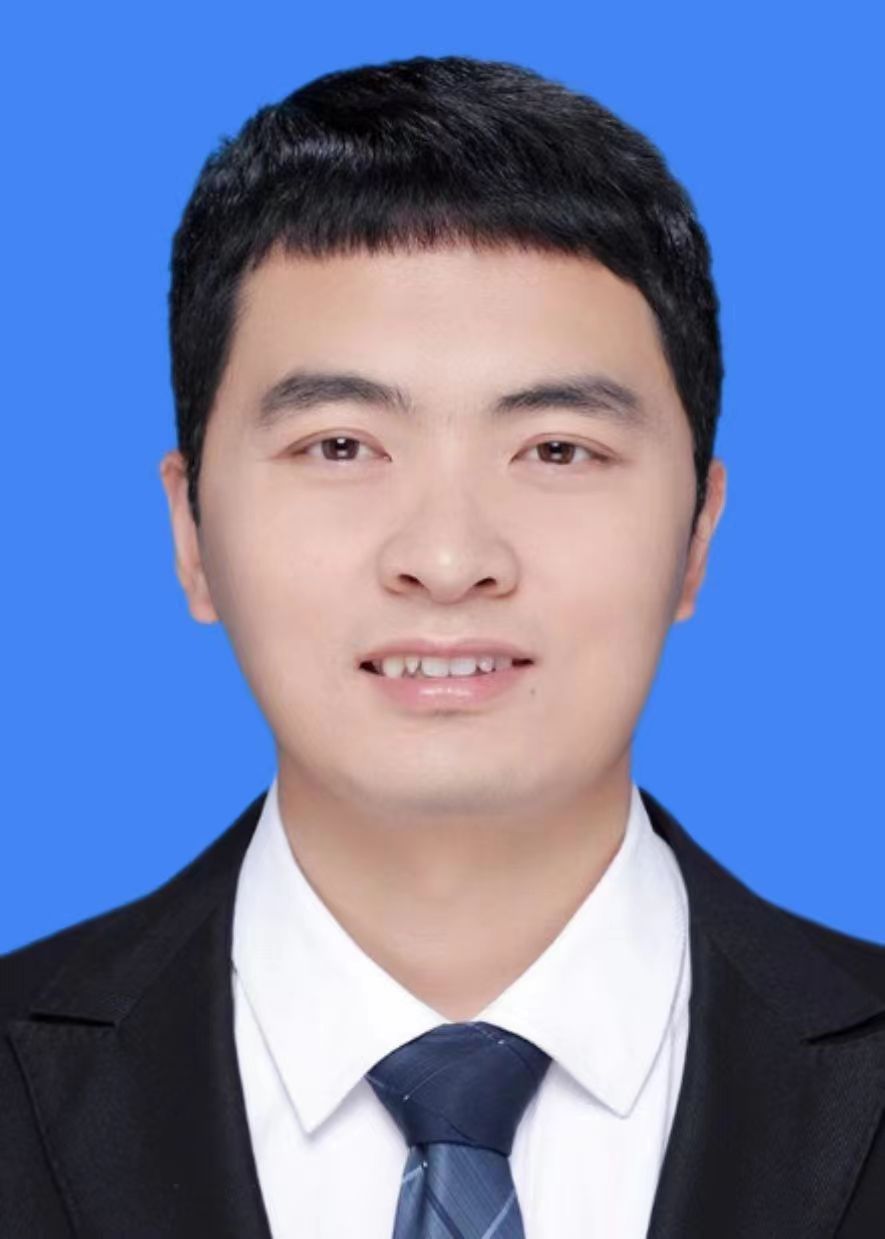}}]{Zhiwei Li}
 , tutor of master students in Beijing University of Chemical Technology. His main research interests include embodied perception, embodied autonomous navigation and localization, and embodied robotic system architecture.
\end{IEEEbiography}

\begin{IEEEbiography}[{\includegraphics[width=1in,height=1.25in,clip,keepaspectratio]{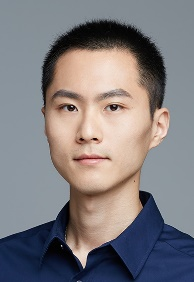}}]{Chen Lv (Senior Member, IEEE)}
is an Associate Professor at School of Mechanical and Aerospace Engineering, and the Cluster Director in Future Mobility Solutions, Nanyang Technological University, Singapore. He joined NTU and founded the Automated Driving and Human-Machine System (AutoMan) Research Lab since June 2018. His research focuses on intelligent vehicles, automated driving, and human-machine systems, where he has published 4 books, over 100 papers, and obtained 12 granted patents. He serves as Associate Editor for IEEE T-ITS, IEEE TVT, and IEEE T-IV. He received many awards and honors, selectively including the IEEE IV Best Workshop/Special Session Paper Award in 2018, Automotive Innovation Best Paper Award in 2020, the winner of Waymo Open Dataset Challenges at CVPR 2021 and 2022, Machines Young Investigator Award, Nanyang Research Award (Young Investigator), SAE Ralph R. Teetor Educational Award, etc.
\end{IEEEbiography}

\begin{IEEEbiography}[{\includegraphics[width=1in,height=1.25in,clip,keepaspectratio]{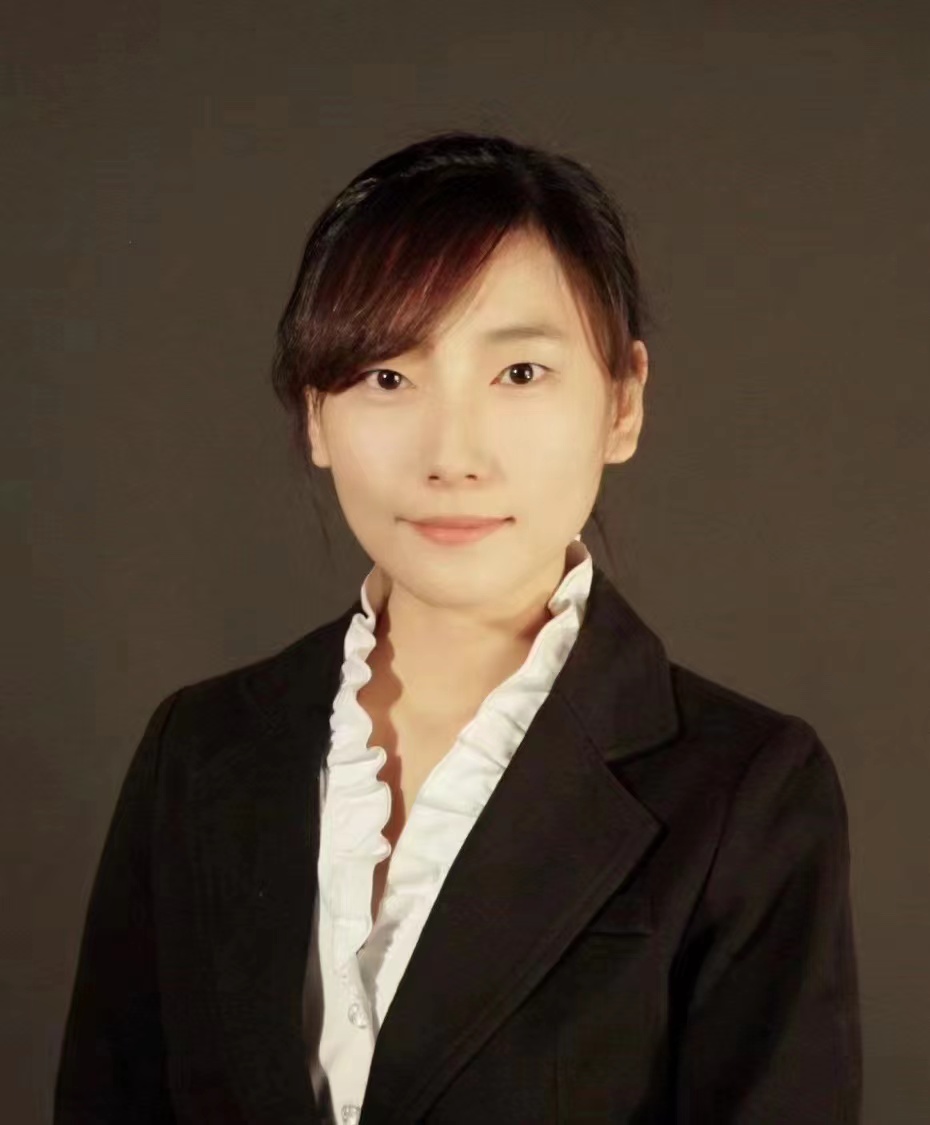}}]{Shanghang Zhang (Member, IEEE)}
is an Assistant Professor at the School of Computer Science, Peking University. Shanghang received her Ph.D. from Carnegie Mellon University and conducted postdoctoral research at the University of California, Berkeley. Committed to research in embodied AI and multimodal large models, she has published over 130 papers on top artificial intelligence journals and conferences. Her works have been cited more than 24,000 times on Google Scholar. She was honored with the Best Paper Award at the AAAI. She has authored the book "Deep Reinforcement Learning" published by Springer Nature, with over 300,000 global e-book downloads. In 2018, she was recognized as an "EECS Rising Star" in the United States and was listed on the "Global AI Chinese Female Young Scholars List" in 2023. Zhang previously won several first place prizes in the International Competitions. She has organized workshops at top international conferences such as NeurIPS and ICML and served as a senior program committee member for AAAI 2022-2026.

\end{IEEEbiography}

\begin{IEEEbiography}[{\includegraphics[width=1in,height=1.25in,clip,keepaspectratio]{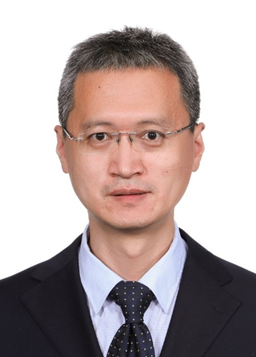}}]{Junqiang Xi (Member, IEEE)}
received the B.S. in Automotive Engineering from Harbin Institute of Technology, Harbin, China, in 1995 and the PhD in Vehicle Engineering from Beijing Institute of Technology, Beijing, China, in 2001. During 2012-2013, he made research as an advanced research scholar in Vehicle Dynamic and Control Laboratory, Ohio State University(OSU), USA. He is currently a Professor with the School of Mechanical Engineering in Beijing Institute of Technology, Beijing, China. His research interests include vehicle dynamic and control, power-train control, mechanics, intelligent vehicles, driver intention recognition, and driver cognitive.
\end{IEEEbiography}

\begin{IEEEbiography}[{\includegraphics[width=1in,height=1.25in,clip,keepaspectratio]{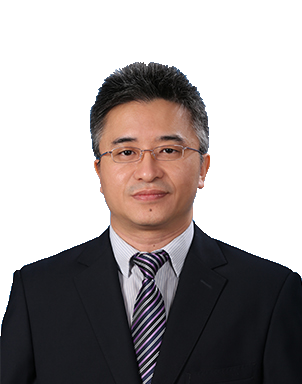}}]{Huaping Liu}
(Fellow, IEEE) received the Ph.D. degree from Tsinghua University, Beijing, China, in 2004. He is currently a Professor with the Department of Computer Science and Technology, at Tsinghua University. His research interests include robot perception and learning. He was a recipient of the National Science Fund for Distinguished Young Scholars. He served as the Area Chair for Robotics Science and Systems for several times. He is a Senior Editor of the International Journal of Robotics Research.
\end{IEEEbiography}

\end{document}